\documentclass[10pt,twocolumn,letterpaper]{article}
\PassOptionsToPackage{numbers, compress}{natbib}
\usepackage{iccv}
\usepackage{times}
\usepackage{epsfig}
\usepackage{graphicx}
\usepackage{amsmath}
\usepackage{amssymb}

\usepackage{wrapfig}    
\usepackage{algorithmic}
\usepackage{algorithm}
\usepackage{multirow}
\usepackage{booktabs}
\usepackage{caption}    
\usepackage{subcaption} 
\usepackage{bm}
\usepackage[table]{xcolor}
\definecolor{mygray}{gray}{0.9}
\definecolor{comment}{RGB}{0, 102, 0}

\newcommand{\norm}[1]{\left\lVert#1\right\rVert}

\usepackage[pagebackref=true,breaklinks=true,letterpaper=true,colorlinks,bookmarks=false]{hyperref}

\iccvfinalcopy 

\def\iccvPaperID{9727} 

\ificcvfinal\fi

\begin{document}

\title{Transporting Causal Mechanisms for Unsupervised Domain Adaptation}

\author{%
 \textbf{Zhongqi Yue}\textsuperscript{1,3} \quad \textbf{Qianru Sun}\textsuperscript{2} \quad \textbf{Xian-Sheng Hua}\textsuperscript{3} \quad \textbf{Hanwang Zhang}\textsuperscript{1}\\
\small \textsuperscript{1}Nanyang Technological University\quad \textsuperscript{2}Singapore Management University \quad \textsuperscript{3}Damo Academy, Alibaba Group\\
\tt\small yuez0003@ntu.edu.sg\quad qianrusun@smu.edu.sg\\
\tt\small\quad xiansheng.hxs@alibaba-inc.com \quad hanwangzhang@ntu.edu.sg\\}

\maketitle
\ificcvfinal\thispagestyle{empty}\fi

\begin{abstract}
Existing Unsupervised Domain Adaptation (UDA) literature adopts the covariate shift and conditional shift assumptions, which essentially encourage models to learn common features across domains. However, due to the lack of supervision in the target domain, they suffer from the semantic loss: the feature will inevitably lose non-discriminative semantics in source domain, which is however discriminative in target domain. We use a causal view---transportability theory~\cite{pearl2014external}---to identify that such loss is in fact a confounding effect, which can only be removed by causal intervention. However, the theoretical solution provided by transportability is far from practical for UDA, because it requires the stratification and representation of the unobserved confounder that is the cause of the domain gap. To this end, we propose a practical solution: Transporting Causal Mechanisms (TCM), to identify the confounder stratum and representations by using the domain-invariant disentangled causal mechanisms, which are discovered in an unsupervised fashion. Our TCM is both theoretically and empirically grounded. Extensive experiments show that TCM achieves state-of-the-art performance on three challenging UDA benchmarks: ImageCLEF-DA, Office-Home, and VisDA-2017. Codes are available at \url{https://github.com/yue-zhongqi/tcm}.
\end{abstract}

\section{Introduction}

\begin{figure}
    \centering
    \includegraphics[width=.98\linewidth]{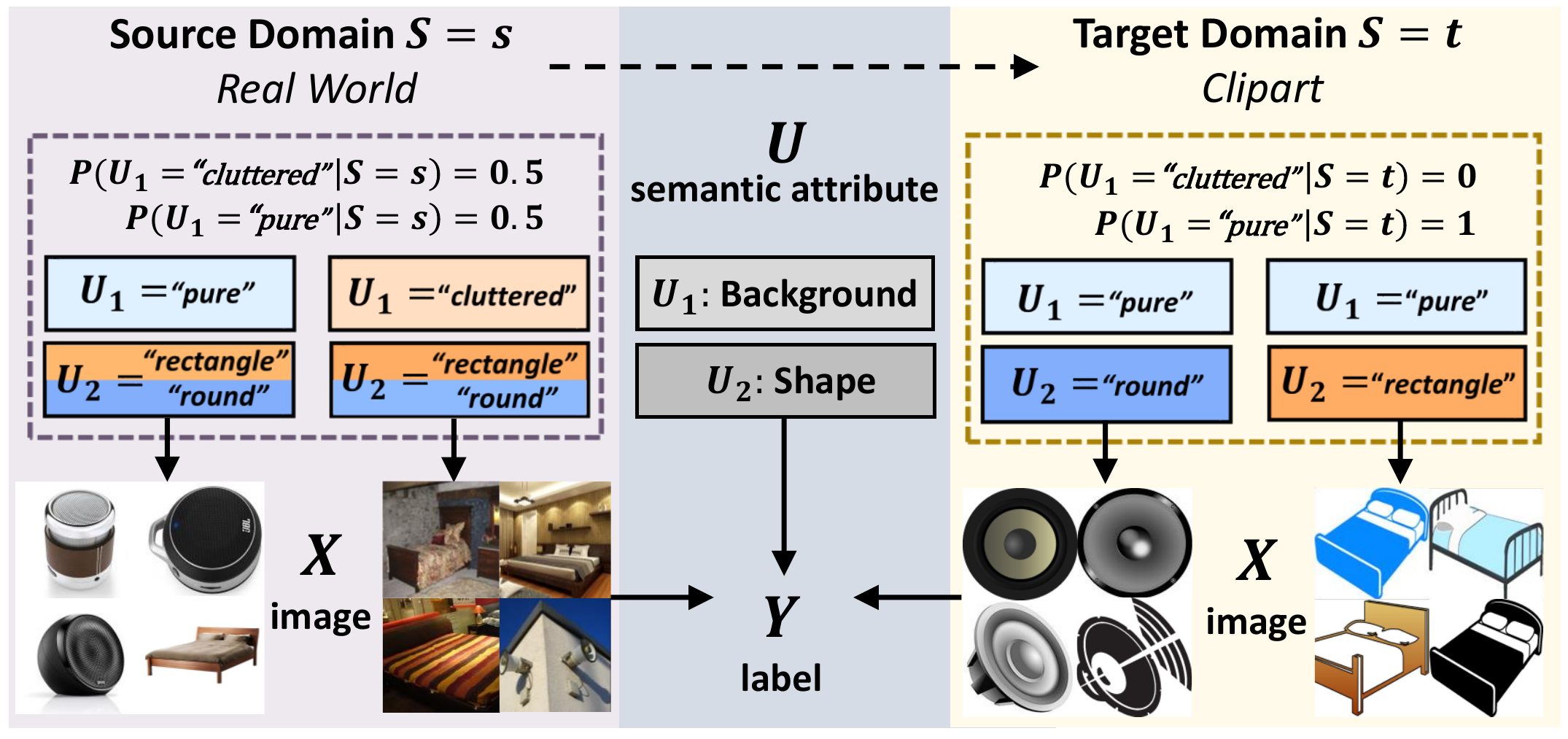}
    \caption{A DA example from source ``Real World'' to target ``Clipart'' domain in the Office-Home benchmark~\cite{venkateswara2017Deep}.}
    \label{fig:1}
    \vspace{-4mm}
\end{figure}

Machine learning is always challenged when transporting training knowledge to testing deployment, which is generally known as Domain Adaptation (DA)~\cite{pan2010domain}. As shown in Figure~\ref{fig:1}, when the \emph{target} domain ($S=t$) is drastically different (\eg, image style) from the \emph{source} domain ($S=s$), deploying a classifier trained in $S=s$ results in poor performance due to the large \emph{domain gap}~\cite{blitzer2007biographies}. To narrow the gap, conventional supervised DA requires a small set of labeled data in $S = t$~\cite{duan2009domain,saenko2010adapting}, which is expensive and sometimes impractical. Therefore, we are interested in a more practical setting: Unsupervised DA (UDA), where we can leverage the abundant unlabelled data in $S=t$~\cite{pan2010domain,long2015learning}. For example, when adapting an autopilot system trained in one country to another, where the street views and road signs are different, one can easily collect unlabelled street images from a camera-equipped vehicle cruise.

Existing UDA literature widely adopts the following two assumptions on the domain gap\footnote{A few early works~\cite{zhang2013domain} also consider target shift of $P(Y)$, which is now mainly discussed in other settings like long-tailed classification~\cite{liu2019large}.}:
1) \emph{Covariate Shift}~\cite{sugiyama2007covariate,bickel2009discriminative}: $P(X|S=s)\neq P(X|S=t)$, where $X$ denotes the samples, \eg, real-world vs. clip-art images; and 
2) \emph{Conditional Shift}~\cite{satpal2007domain,long2013transfer}: $P(Y|X,S=s)\neq P(Y|X,S=t)$, where $Y$ denotes the labels, \eg, in clip-art domain, the ``pure background'' feature extracted from $X$ is no longer a strong visual cue for $Y=$ ``speaker'' as in real-world domain. To turn ``$\neq$'' into ``$=$'', almost all existing UDA solutions rely on  learning invariant (or common) features in both source and target domains~\cite{sugiyama2007covariate, hoffman2018cycada, long2015learning}. Unfortunately, due to the lack of supervision in target domain, it is challenging to capture such domain invariance. 

For example, in Figure~\ref{fig:1}, when $S = s$, the distribution of ``Background = pure'' and ``Background = cluttered'' is already informative for ``speaker'' and ``bed'', a classifier may recklessly tend to downplay the ``Shape'' features (or attributes) as they are not as ``invariant'' as ``Background''; but, it does not generalize to $S = t$, where all clip-art images have ``Background = pure'' that is no longer discriminative. 
\begin{wrapfigure}{r}{0.14\textwidth}
    \centering
    \includegraphics[width=.14\textwidth]{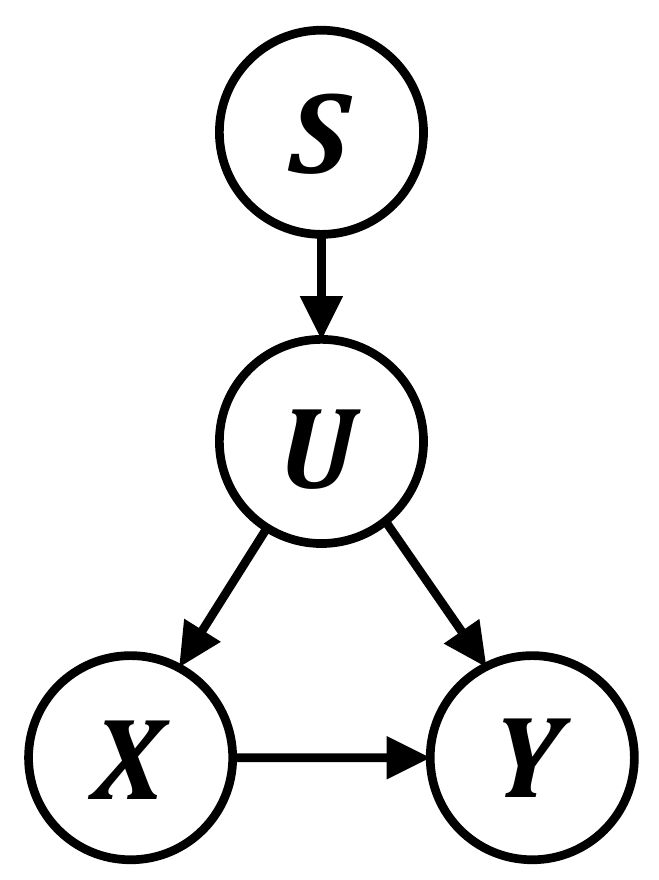}
    \caption{Causal graph of domain adaptation.}
    \label{fig:2}
\end{wrapfigure}
A popular approach to make up for the above loss of ``Shape'' is by imposing unsupervised reconstruction~\cite{bousmalis2016domain}---any feature loss hurts reconstruction. However, it is well-known
that discrimination and reconstruction are adversarial objectives, leaving the finding of their ever-elusive trade-off parameter \emph{per se} an open problem~\cite{chen2018zero,ghifary2016deep,kondor2018generalization}.

To systematically understand how domain gap causes the feature loss, we propose to use the \emph{transportability theory}~\cite{pearl2014external} to replace the above ``shift'' assumptions that overlook the explicit role of semantic attributes. Before we introduce the theory, we first abstract the DA problem in Figure~\ref{fig:1} into the causal graph in Figure~\ref{fig:2}. We assume that the classification task $X\to Y$ is also affected by the unobserved semantic feature $U$ (\eg, shape and background), where $U\to X$ denotes the generation of pixel-level image samples $X$ and $U\to Y$ denotes the 
definition process of semantic class. Note that these causalities have been already shown valid in their respective areas, \eg, $U\to X$ and $U\to Y$ can be image~\cite{goodfellow2014generative} and language generation~\cite{brown2020language}, respectively. In particular, the introduction of the domain selection variable $S$ reveals that the fundamental reasons of the covariate shift and conditional shift are both due to $P(U|S=s)\neq P(U|S=t)$.

Note that the domain-aware $U$ is the \emph{confounder} that prevents the model from learning the domain-invariant causality $X\to Y$. It has been theoretically proven that the confounding effect cannot be eliminated by statistical learning without causal intervention~\cite{pearl2009causality}. Fortunately, the transportability theory offers a principled DA solution based on the causal intervention in Figure~\ref{fig:2}:
\begin{equation}
    P(Y|do(X),S)
    =\sum\nolimits_{u} P(Y|X,U=u)P(U=u|S).
    \label{eq:transportability}
\end{equation}
Note that the goal of DA is achieved by causal intervention $P(Y|do(X), S)$ using the $do$-operator~\cite{pearl2016causal}. To appreciate the merits of the calculus on the right-hand side of Eq.~\eqref{eq:transportability}, we need to understand the following two points. First, $P(Y|X, U)$ is domain-agnostic as $S$ is separated from $X$ and $Y$ given $U = u$~\cite{pearl2009causality}, thus $P(Y|X, U)$ generalizes to $S = t$ in testing even if it is trained on $S = s$. Second, every stratum of $U$ is fairly adjusted subject to the domain prior $P(U|S)$, \ie, forcing the model to respect ``shape'' as it is the only ``invariance'' that distinguishes between ``speaker'' and ``bed'' in controlled ``cluttered'' or ``pure'' backgrounds. Therefore, the semantic loss is eliminated.

However, Eq.~\eqref{eq:transportability} is purely declarative and far from practical, because it is still unknown how to implement the stratification and representation of the \emph{unobserved} $U$, not mentioning to transport it across domains. To this end, we propose a novel approach to make Eq.~\eqref{eq:transportability} practical in UDA:

\noindent\textbf{Identify the number of $U$ by disentangled causal mechanisms}. In practice, the number of $U$ can be large due to the combinations of multiple attributes, making Eq.~\eqref{eq:transportability} computationally prohibitive (\eg, many network passes). Fortunately, if the attributes are disentangled, we can stratify $U$ according to the much smaller number of disentangled attributes, as the effect of each feature is independent with each other~\cite{higgins2018towards}, \eg, $P(Y|X, U = (u_1, u_2)) \propto P(Y|X, u_1 ) + P(Y|X,u_2)$ if $u_1$ and $u_2$ are disentangled. As detailed in Section~\ref{sec:3.1}, this motivates us to discover a small number of Disentangled Causal Mechanisms (DCMs) in unsupervised fashion~\cite{parascandolo2018learning}, each of which corresponds to a feature-specific
intervention
between $X|S=s$ and $X|S=t$,
\eg, mapping a real-world ``speaker'' image to its clip-part counterpart by changing ``Background'' from ``cluttered'' to ``pure''.

\noindent\textbf{Represent $U$ by proxy variables}. Yet, DCMs do not provide vector representations of $U$, leaving difficulties in implementing Eq.~\eqref{eq:transportability} as neural networks.
In Section~\ref{sec:3.2}, we show that the transformed output $\hat{X}$ of each DCM taking $X$ as input can be viewed as a proxy of $U$~\cite{miao2018identifying}, who provides a theoretical guarantee
that we can replace the \emph{unobserved} $U$ with the \emph{observed} $\hat{X}$ to make Eq.~\eqref{eq:transportability} computable.

Overall, instead of transporting the abstract and unobserved $U$ in the original transportability theory of Eq.~\eqref{eq:transportability}, we transport the concrete and learnable disentangled causal mechanisms, which generate the observed proxy of $U$. Therefore, we term the approach as \textbf{Transporting Casual Mechanisms (TCM)} for UDA. Through extensive experiments, our approach achieves state-of-the-art performance on UDA benchmarks: 70.7\% on Office-Home~\cite{venkateswara2017Deep}, 90.5\% on ImageCLEF-DA~\cite{imageclef2014} and 75.8\% on VisDA-2017~\cite{peng2017visda}. Specifically, we show that learning disentangled mechanisms and capturing the causal effect through proxy variables are the keys to improve the performance, which validates the effectiveness of our approach.

\section{Related Work}
Existing UDA works fall into the three categories:
\textbf{1) Sample Reweighing}~\cite{sugiyama2007covariate,chen2018re}. This line-up adopts the covariate shift assumption. It first models $P(X|S=s)$ and $P(X|S=t)$. When minimizing the classification loss, each sample in $S=s$ is assigned an importance weight $P(X|S=t)/P(X|S=s)$, \ie, a target-like sample has a larger weight, which effectively encourages $P(X|S=s)=P(X|S=t)$.
\textbf{2) Domain Mapping}~\cite{hoffman2018cycada,murez2018image}. This approach focuses on the conditional shift assumption. It first learns a mapping function that transforms samples from $S=s$ to $S=t$ through unpaired image-to-image translation techniques such as CycleGAN~\cite{zhu2017unpaired}. Then, the transformed source domain samples are used to train a classifier.
\textbf{3) Invariant Feature Learning}. This is the most popular approach in recent literature~\cite{ganin2016domain,long2018conditional,xu2020reliable}. It maps the samples in $S=s$ and $S=t$ to a common feature space, denoted as domain $S=c$, where the source and target domain samples are indistinguishable. Some works~\cite{pan2010domain,long2015learning} adopted the covariate shift assumption, and aim to minimize the differences in $P(X|S=c)$ using distance measures like Maximum Mean Discrepancy~\cite{tzeng2014deep,long2014transfer} or using adversarial training to fool a domain discriminator~\cite{ganin2016domain,bousmalis2016domain}. Others used the conditional shift assumption and aim to align $P(Y|X,S=c)$~\cite{long2013transfer,long2018conditional}. As the target domain samples have no labels, their $P(Y|X,S=c)$ is either estimated through clustering~\cite{kang2019contrastive,zhang2020label} or using a classifier trained with the source domain samples~\cite{kuniaki2017asymmetric,zhang2019domain}.

They all aim to make the source and target domain alike while learning a classifier with the labelled data in the source domain, leading to the semantic loss. In fact, there are existing works that attempt to \textbf{Alleviate Semantic Loss} by learning to capture $U$ in unsupervised fashion: One line exploits that $U$ generates $X$ (via $U\to X$) and learns a latent variable to reconstruct $X$~\cite{bousmalis2016domain,ghifary2016deep}. The other line~\cite{cai2019llearning,cui2020hda} exploits $S\to U$, and aims to learn a domain-specific representation for each $S=s, S=t$. However, this leads to an ever-illusive trade-off between the adversarial discrimination and reconstruction objectives. Our approach aims to fundamentally eliminate the semantic loss by providing a practical implementation of the transportability theory~\cite{pearl2014external} based on causal intervention~\cite{pearl2016causal}.
\section{Approach}

Though Eq.~\eqref{eq:transportability} provides a principled solution for DA, it is impractical due to the unobserved  $U$. To this end, the proposed TCM is a two-stage approach to make it practical. The first stage identifies the $U$ stratification by discovering Disentangled Causal Mechanisms (DCMs) ( Secton~\ref{sec:3.1}) and the second stage represents each stratification of $U$ with the proxy variable generated from the discovered DCMs (Section~\ref{sec:3.2}). The training and inference procedures are summarized in Algorithm~\ref{alg:1} and~\ref{alg:2}, respectively.

\renewcommand{\algorithmicforall}{\textbf{for each}}
\begin{algorithm}[!h]
  \caption{Two-stage Training of TCM}
  \label{alg:1}
\begin{algorithmic}[1]

\STATE {\bfseries Input:} Labelled domain $S=s$, unlabelled domain $S=t$, pre-trained backbone parameters $\beta$
\STATE {\bfseries Output:} DCMs $\{(M_i,M_i^{-1})\}_{i=1}^{k}$, fine-tuned backbone parameters $\beta$ and linear functions parameters $\omega$
  
\STATE Randomly initialize $\{(M_i,M_i^{-1})\}_{i=1}^{k},\omega$, VAE parameters $\theta$, discriminator parameters $\gamma$ (see Eq.~\eqref{eq:proxy_objective})
\REPEAT
    \STATE \textcolor{blue}{// See Section 3.1 for details}
    \STATE Sample $\mathbf{x}$ randomly from $S=s$ or $S=t$
    \STATE Calculate $\mathcal{L}_{CycleGAN}^i$ for $i\in\{1,\ldots,k\}$
    \STATE Update $\{(M_i,M_i^{-1})\}_{i=1}^{k}$ with Eq.~\eqref{eq:dcm_objective}
\UNTIL{convergence}

\REPEAT
    \STATE \textcolor{comment}{// See Section 3.2 for details}
    \STATE Sample $\mathbf{x}_s, y_s$ from $S=s$, $\mathbf{x}_t$ from $S=t$
    \STATE Obtain $\hat{\mathcal{X}}_s=\{M_i(\mathbf{x}_s)\}_{i=1}^k$,$\hat{\mathcal{X}}_t=\{M_i^{-1}(\mathbf{x}_t)\}_{i=1}^k$
    \STATE Calculate $h_y(\mathbf{x}_s,\hat{\mathbf{x}}), \forall \hat{\mathbf{x}}\in \hat{\mathcal{X}_s}$ with Eq.~\eqref{eq:h_y}
    \STATE Obtain $P(Y|do(X=\mathbf{x}_s),S=s)$ with Eq.~\eqref{eq:practical_transport}
    \STATE Update $\omega,\beta,\theta,\gamma$ with Eq.~\eqref{eq:proxy_objective}
\UNTIL{convergence}
\end{algorithmic}
\end{algorithm}
\begin{algorithm}[!h]
  \caption{Inference with TCM}
  \label{alg:2}
\begin{algorithmic}[1]

\STATE {\bfseries Input:} $\mathbf{x}_t$, $P(\hat{X}|S=t)$, DCMs $\{M_i^{-1}\}_{i=1}^{k}$, backbone parameters $\beta$ and linear functions parameters $\omega$
\STATE {\bfseries Output:} Predicted label $y_t$

    \STATE Obtain $\hat{\mathcal{X}}_t=\{M_i^{-1}(\mathbf{x}_t)\}_{i=1}^k$
    \STATE Calculate $h_y(\mathbf{x}_t,\hat{\mathbf{x}}), \forall \hat{\mathbf{x}}\in \hat{\mathcal{X}}_t$ with Eq.~\eqref{eq:h_y}
    \STATE $y_t\leftarrow \mathrm{argmax}\;P(Y|do(X=\mathbf{x}_t),S=t)$ with Eq.~\eqref{eq:practical_transport}

\end{algorithmic}
\end{algorithm}

\subsection{Disentangled Causal Mechanisms Discovery}
\label{sec:3.1}

Physicists believe that real-world observations are the outcome of the combination of independent physical laws. In causal inference~\cite{higgins2018towards,suter2019robustly}, we denote these laws as disentangled generative factors, such as shape, color, and position. For example,
in our causal graph of Figure~\ref{fig:2}, if the semantic attribute $U$ can be disentangled into generative causal factors $(U_1,\ldots,U_k)$, each one will independently contribute to the observation: $P(Y|X)\propto\sum_{i=1}^k P(Y|X,U_i)$. Even though  $|U|$ is large, \ie, the number of sum in Eq.~\eqref{eq:transportability} is expensive, the disentanglement can still significantly reduce the number to a small $k$.

\begin{figure}
    \centering
    \includegraphics[width=1\linewidth]{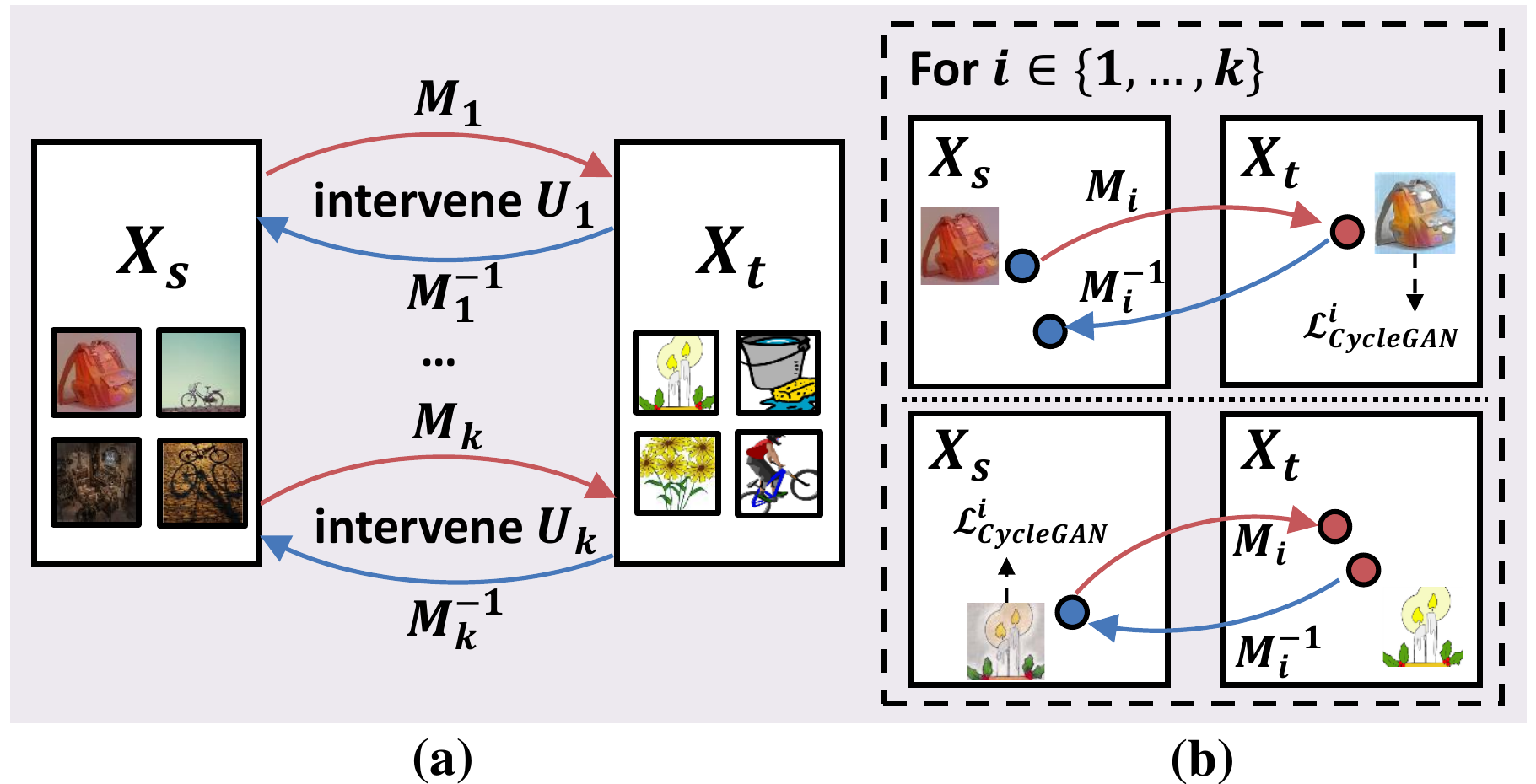}
    \caption{(a) Our DCMs $\{(M_i,M_i^{-1})\}_{i=1}^k$, where each $M_i:X_s\to X_t$ and $M_i^{-1}:X_t \to X_s$ correspond to intervention on the disentangled attribute $U_i$. (b) CycleGAN~\cite{zhu2017unpaired} loss $\mathcal{L}_{CycleGAN}^i$ on each $(M_i,M_i^{-1})$.}
    \label{fig:3}
    \vspace{-4mm}
\end{figure}

However, learning a disentangled representation of $U=(U_1,\ldots,U_k)$ without supervision is challenging~\cite{locatello2019challenging}. Fortunately, we can observe the outcomes of $U$: the source domain samples $X_s$ and the target domain samples $X_t$ generated from $P(X|U)$.
This motivate us to discover $k$ pairs of end-to-end functions $\{(M_i,M_i^{-1})\}_{i=1}^k$ in unsupervised fashion, where $M_i:X_s\to X_t$ and $M_i^{-1}:X_t \to X_s$, as shown in Figure~\ref{fig:3} (a). Each $(M_i,M_i^{-1})$ corresponds to a counterfactual mapping that transforms a sample to the counterpart domain by intervening a disentangled factor $U_i$, \ie, modifying $U_i$ \wrt the domain shift while fixing the values of other factors $U_{j\neq i}$---this is essentially the definition of independent causal mechanisms~\cite{parascandolo2018learning}, each of which corresponds to a disentangled causal factor. Hence we refer to these mapping functions as DCMs.

We first make a strong assumption that the one-to-one correspondence between $(M_i,M_i^{-1})$ and $U_i$ has been already established, then we relax this assumption by removing the correspondence for the proposed practical solution. Without loss of generality, we consider a source domain sample $\mathbf{x}_s$. For each $M_i:X_s\to X_t$, we obtain a transformed sample $M_i(\mathbf{x}_s)$, corresponding to the interventional outcome of $U_i$ drawn from $P(U_i|S=s)$ to $P(U_i|S=t)$. To ensure that the interventions are disentangled, we use the Counterfactual Faithfulness theorem~\cite{besserve2020counterfactual,yue2021counterfactual}, which guarantees that $M_i$ is a disentangled intervention if and only if $M_i(\mathbf{x}_s)\sim P(X_t)$ (proof in Appendix). Hence, as shown in Figure~\ref{fig:3} (b), we use the CycleGAN loss~\cite{zhu2017unpaired} denoted as $\mathcal{L}_{CycleGAN}^i$ to make the transformed samples from \emph{each} $M_i:X_s\to X_t$ indistinguishable from the real samples in $S=t$. Likewise, we apply the CycleGAN loss to each $M_i^{-1}: X_t \to X_s$ such that the transformed samples are close to real samples in $S=s$.

\begin{figure}[t]
    \centering
    \includegraphics[width=.85\linewidth]{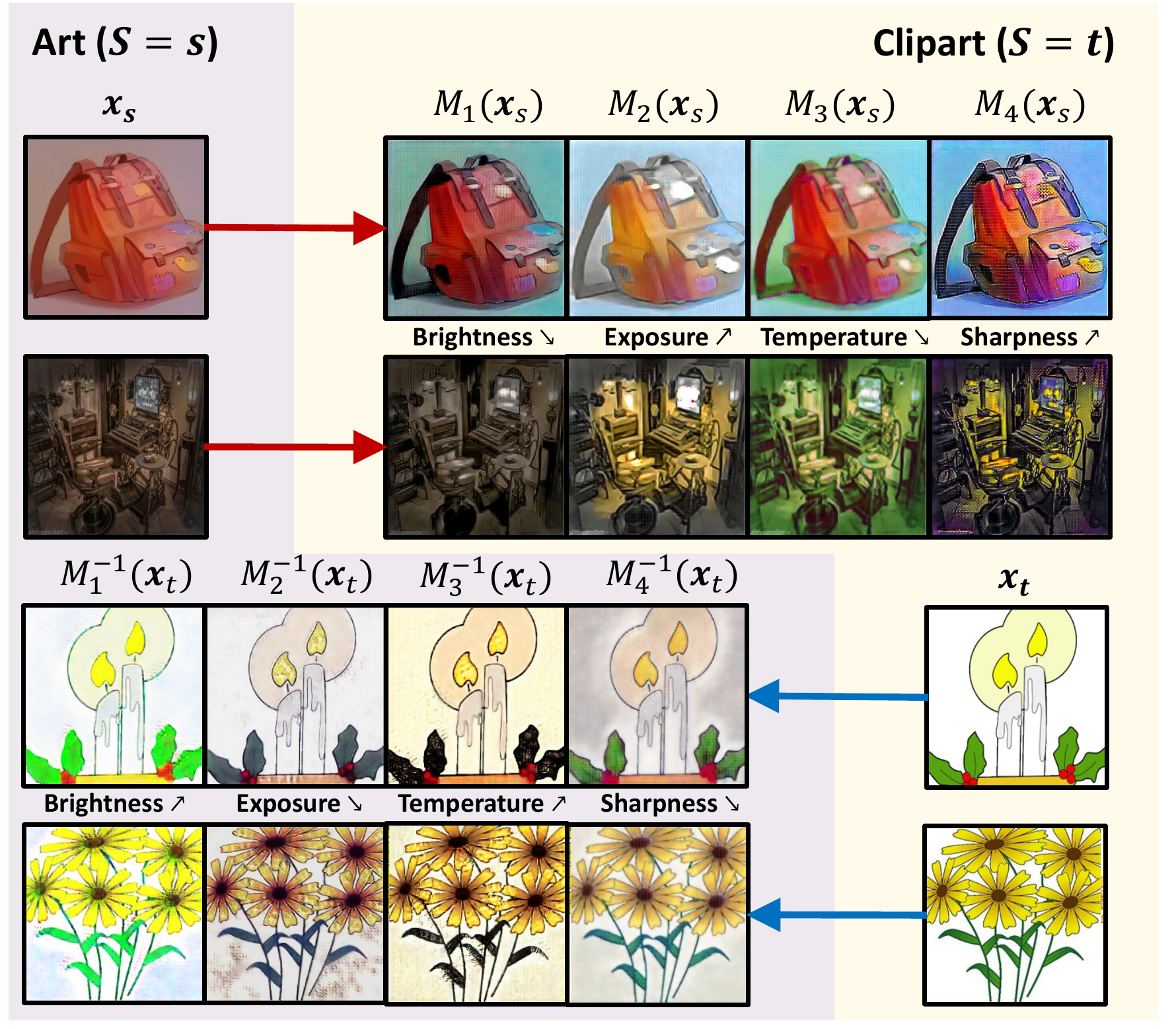}
    \caption{Transformation between ``Art'' and ``Clipart'' domain with 4 trained DCMs $\{(M_i,M_i^{-1})\}_{i=1}^4$, where each pair of mapping functions specializes in brightness, exposure, color temperature and sharpness, respectively.}
    \label{fig:4}
\end{figure}

Now we relax the above one-to-one correspondence assumption. We begin with a sufficient condition (proof in Appendix): \emph{if $(M_i,M_i^{-1})$ intervenes $U_i$, the $i$-th mapping function outputs the counterfactual faithful generation}, \ie, $i = \mathop{\mathrm{argmin}}_{j\in\{1,\ldots,k\}} \mathcal{L}_{CycleGAN}^j$. To ``guess'' the one-to-one correspondence, we adopt a practical method by using the necessary conclusion: if $i = \mathop{\mathrm{argmin}}_{j\in\{1,\ldots,k\}} \mathcal{L}_{CycleGAN}^j$, then $(M_i,M_i^{-1})$ corresponds to $U_i$ intervention. Specifically, training samples are fed into all $(M_i,M_i^{-1})$ in parallel to compute $\mathcal{L}_{CycleGAN}^i$ for each pair. Only the winning pair with the smallest loss is updated. The objective is given by:
\begin{equation}
    \begin{aligned}
        &\mathop{\mathrm{min}}_{(M_i,M_i^{-1})} \mathcal{L}_{CycleGAN}^i,\\
        \mathrm{where} \; &i = \mathop{\mathrm{argmin}}_{j\in\{1,\ldots,k\}} \mathcal{L}_{CycleGAN}^j,
    \end{aligned}
    \label{eq:dcm_objective}
\end{equation}
The functions (\eg, $M_i$) in the optimization objective denote their parameters for simplicity.
Note that this is not sufficient, hence our approach has limitations. Still, this practice has been empirically justified in~\cite{parascandolo2018learning}, and we leave a more theoretically complete approach as future work.

After training, we obtain $k$ DCMs $\{(M_i,M_i^{-1})\}_{i=1}^k$, where the $i$-th pair corresponds to $U_i$. Hence we identify the number of $U$ as $k$. Figure~\ref{fig:4} shows an example with $k=4$, where the 4 DCMs correspond to ``Brightness'', ``Exposure'', ``Temperature'' and ``Sharpness'', respectively\footnote{Note that our DCM learning is orthogonal to the GAN models used. One can use more advanced GAN models to discover complex mechanisms such as shape or viewpoint changes.}.

\subsection{Representing $U$ by Proxy Variables}
\label{sec:3.2}

\begin{wrapfigure}{r}{0.17\textwidth}
    \centering
    \includegraphics[width=.17\textwidth]{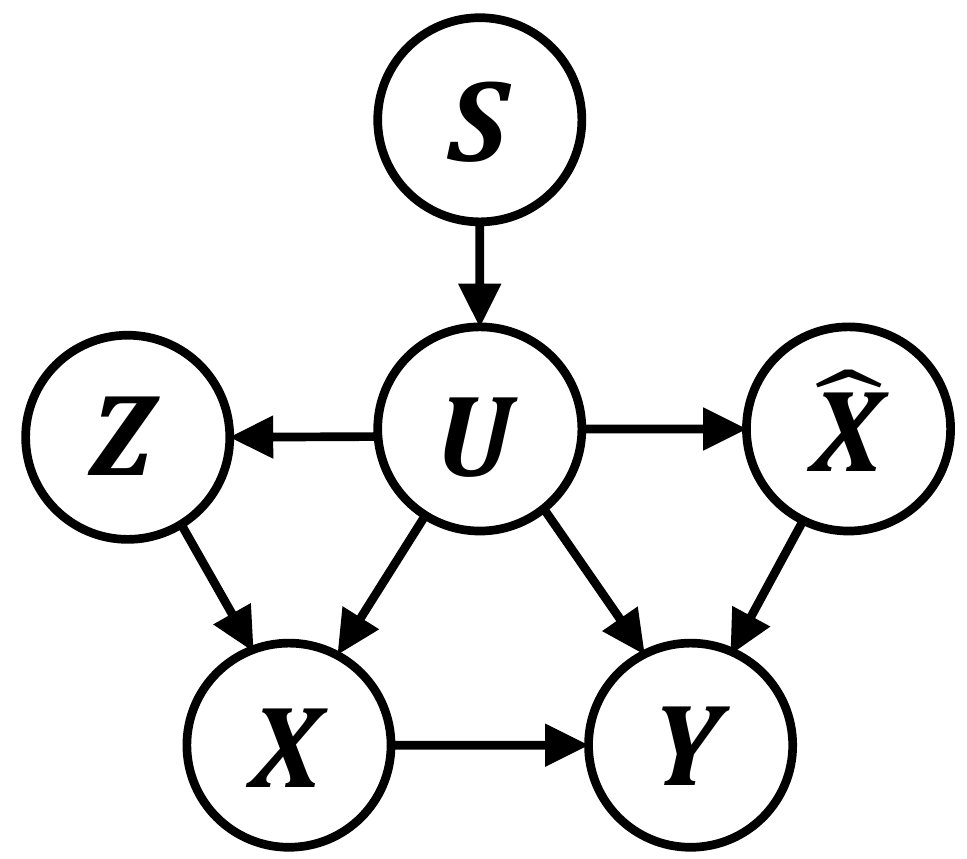}
    \caption{Causal graph with the proxy $\hat{X}$ and $Z$.}
    \label{fig:5}
    \vspace{-3mm}
\end{wrapfigure}

While the learned DCMs identify the number of $U$, they do not provide a vector representation of $U$. Hence $P(U|S)$ and $P(Y|X,U)$ in Eq.~\eqref{eq:transportability} are still unobserved.
Fortunately, $U$ generates two \emph{observed} variables $\hat{X}$ and $Z$ as shown in Figure~\ref{fig:5}, which are called \emph{proxy variables}~\cite{miao2018identifying} of $U$ and make $P(U|S)$ and $P(Y|X,U)$ estimable. We will first explain $\hat{X}$ and $Z$ as well as their related causal links.

\noindent$\mathbf{\hat{X},U \to \hat{X} \to Y}$. $\hat{X}$ represents the DCMs outputs, \ie, for a sample $\mathbf{x}_s$ in $S=s$, $\hat{X}|S=s$ takes value from $\hat{\mathcal{X}}_s = \{M_i(\mathbf{x}_s)\}_{i=1}^k$; and for a sample $\mathbf{x}_t$ in $S=t$, $\hat{X}|S=t$ takes value from $\hat{\mathcal{X}}_t = \{M_i^{-1}(\mathbf{x}_t)\}_{i=1}^k$. As detailed in Appendix, $M_i,M_i^{-1}$ generates counterfactuals in the other domain by fixing $U_{j\neq i}$ and intervening $U_i$, \ie, $\hat{X}$ is generated from $U$. Hence $U\to \hat{X}$ is justified. Moreover, as the other domain is also labelled, $\hat{X} \to Y$ denotes the predictive effects from the generated counterfactuals.

\noindent$\mathbf{Z,U\to Z \to X}$. $Z$ is a latent variable encoded from $X$ by a VAE~\cite{kingma2014ICLR}. The link $Z \to X$ is because the latent variable is trained to reconstruct $X$. Furthermore, the latent variable of VAE is shown to capture some information of the underlying $U$~\cite{suter2019robustly}, justifying $U\to Z$.

Note that the networks to obtain $\hat{X}$ and $Z$ are trained in an unsupervised fashion. Hence $\hat{X},Z$ are observed in both $S=s$ and $S=t$. Under this causal graph, we have a theoretically grounded solution to Eq.~\eqref{eq:transportability} using the theorem below (proof in Appendix as corollary to~\cite{miao2018identifying}).

\noindent\textbf{Theorem (Proxy Function)}. \textit{Under the causal graph in Figure~\ref{fig:5}, any solution $h_y(X,\hat{X})$ to Eq.~\eqref{eq:proxy1} satisfies Eq.~\eqref{eq:proxy2}.}
\begin{equation}
    P(Y|Z,X,S)=\sum_{\mathbf{\hat{x}}} h_y(X,\mathbf{\hat{x}})P(\hat{X}=\mathbf{\hat{x}}|Z,X,S)
    \label{eq:proxy1}
\end{equation}
\begin{equation}
    P(Y|X,U)=\sum_{\mathbf{\hat{x}}} h_y(X, \mathbf{\hat{x}})P(\hat{X}=\mathbf{\hat{x}}|U).
    \label{eq:proxy2}
\end{equation}

Notice that we can set $S=s$ in Eq.~\eqref{eq:proxy1} and use the labelled data in $S=s$ to learn $h_y$ (details given below). We prove in the Appendix that $h_y$ is invariant across domains. This leads to the following inference strategy.

\subsubsection{Inference}

Taking expectation over $U|S=t$ on both sides of Eq.~\eqref{eq:proxy2} leads to a practical solution to Eq.~\eqref{eq:transportability}:
\begin{equation}
    P(Y|do(X),S=t)=\sum_{\mathbf{\hat{x}} \in \hat{\mathcal{X}}_t} h_y(X,\mathbf{\hat{x}}) P(\hat{X}=\mathbf{\hat{x}}|S=t),
    \label{eq:practical_transport}
\end{equation}
where $P(\hat{X}|S=t)$ is estimable as $\hat{X}$ is observed in $S=t$, and $h_y$ is trained in $S=s$ as given below.

\subsubsection{Learning $h_y$}
We detail the training of $h_y$, including the representation of its inputs $X$ and $\hat{X}$, the function forms of the terms in Eq.~\eqref{eq:proxy1} and the training objective.

\noindent\textbf{Representation of $X,\hat{X}$}. When $X$ and $\hat{X}$ are images, directly evaluating Eq.~\eqref{eq:practical_transport} in image space can be computationally expensive, not to mention the tendency for severe over-fitting. Therefore, we follow the common practice in UDA~\cite{long2018conditional,kang2019contrastive} to introduce a pre-trained feature extractor backbone (\eg, ResNet~\cite{he2016deep}). Hereinafter, $X,\hat{X}$ denote their respective $n$-d features. Note that $Z$ of $l$ dimensions is encoded from $X$'s feature form ($l < n$).

\noindent\textbf{Function Forms}. We adopt an isotropic Gaussian distribution for $P(\hat{X}|S)$.
We implement $P(Y|Z,X,S=s),P(\hat{X}|Z,X,S=s)$ in Eq.~\eqref{eq:proxy1} with the following linear model $f_y(Z,X)$ and $f_{\hat{x}}(Z,X)$, respectively:
\begin{equation}
    \begin{split}
        &f_y(Z,X)=\mathbf{W}_1 Z + \mathbf{W}_2 X + \mathbf{b}_1 \\
        &f_{\hat{x}}(Z,X)=\mathbf{W}_3 Z + \mathbf{W}_4 X + \mathbf{b}_2,
    \end{split}
    \label{eq:linear}
\end{equation}
where $f_y:(Z,X)\to \mathbb{R}^c$ produces logits for $c$ classes,
$f_{\hat{x}}:(Z,X)\to \mathbb{R}^n$ predicts $\hat{X}$,
$\mathbf{W}_1 \in \mathbb{R}^{c\times l}$,
$\mathbf{W}_2 \in \mathbb{R}^{c\times n}$,
$\mathbf{W}_3 \in \mathbb{R}^{n\times l}$,
$\mathbf{W}_4 \in \mathbb{R}^{n\times n}$,
$\mathbf{b}_1 \in \mathbb{R}^c$ and
$\mathbf{b}_2 \in \mathbb{R}^n$.
With this function form, we can solve $h_y:(X,\hat{X})\to \mathbb{R}^c$ from Eq.~\eqref{eq:proxy1} as (derivation in Appendix):
\begin{equation}
    \begin{split}
        h_y(X,\hat{X})&=\mathbf{b}_1-\mathbf{W}_1\mathbf{W}_3^+\mathbf{b}_2 + \mathbf{W}_1\mathbf{W}_3^+ \hat{X} \\
        &+ (\mathbf{W}_2-\mathbf{W}_1\mathbf{W}_3^+\mathbf{W}_4)X,
        \label{eq:h_y}
    \end{split}
\end{equation}
where $(\cdot)^+$ denotes pseudo-inverse.

\noindent\textbf{Overall Objective}. The training objective is given by:
\begin{equation}
    \mathop{\mathrm{min}}_{\omega,\beta,\theta} (\mathcal{L}_c + \mathcal{L}_{v}) + \mathop{\mathrm{min}}_{\beta} \mathop{\mathrm{max}}_{\gamma} \alpha \mathcal{L}_p,
    \label{eq:proxy_objective}
\end{equation}
where $\omega$ denotes the parameters of the linear functions $f_y$ and $f_{\hat{x}}$,
$\beta$ denotes the parameters of the backbone,
$\theta$ denotes the parameters of the VAE,
$\mathcal{L}_c$ denotes the sum of Cross-Entropy loss to train $f_y$ and the mean squared error loss to train $f_{\hat{x}}$ (see Eq.~\eqref{eq:linear}).
$\mathcal{L}_v$ is the VAE loss,
$\mathcal{L}_p$ is the proxy loss,
$\gamma$ denotes the parameters of the discriminators used in $\mathcal{L}_p$,
and $\alpha$ is a trade-off parameter.

\noindent\textbf{VAE Loss}. $\mathcal{L}_v$ is used to train the VAE, which contains an encoder $Q_\theta(Z|X)$ and a decoder $P_\theta(X|Z)$. Given a feature $\mathbf{x}_s$ in $S=s$, $\mathcal{L}_v$ is given by:
\begin{equation}
\begin{split}
    \mathcal{L}_v &= -\mathbb{E}_{Q_\theta(Z|X=\mathbf{x}_s)} [P_\theta(X=\mathbf{x}_s|Z)]\\ 
    &+ D_{KL}(Q_\theta(Z|X=\mathbf{x}_s) \mid\mid P(Z)),
\end{split}
\end{equation}
where $D_{KL}$ denotes KL-divergence and $P(Z)$ is set to the isotropic Gaussian distribution. Note that we only need to learn a VAE in $S=s$, as $h_y(X,\hat{X})$ is domain-agnostic.

\noindent\textbf{Proxy Loss}. In practice, the generated images from the DCMs $\{(M_i,M_i^{-1})\}_{i=1}^k$ may contain artifacts that contaminate the feature outputs from the backbone~\cite{lu2017safetynet}, making $\hat{X}$ as feature no longer resembles the features in the counterpart domain.
Hence we regularize the backbone with a proxy loss to enforce feature-level resemblance between the samples transformed by the DCMs and the samples in the counterpart domain. Given feature $\mathbf{x}_s$ in $S=s$, $\mathbf{x}_t$ in $S=t$ and their corresponding DCMs outputs sets $\hat{\mathcal{X}_s},\hat{\mathcal{X}_t}$ (as $n$-d features), the loss is given by:
\begin{equation}
    \begin{split}
        \mathcal{L}_p &= \mathrm{log} D_s(\mathbf{x}_s)+ \frac{1}{k}\sum_{\mathbf{\hat{x}_s} \in \hat{\mathcal{X}_s}} \mathrm{log}(1-D_t(\mathbf{\hat{x}_s})) \\
        & + \mathrm{log} D_t(\mathbf{x}_t)+\frac{1}{k}\sum_{\mathbf{\hat{x}_t} \in \hat{\mathcal{X}_t}} \mathrm{log}(1-D_s(\mathbf{\hat{x}}_t)),
    \end{split}
    \label{eq:la}
\end{equation}
where $D_s,D_t$ are discriminators parameterized by $\gamma$ that return a large value for features of real samples.
Through min-max adversarial training, generated sample features $\hat{X}$ become similar to the sample features in the counterpart domain and fulfill the role as a proxy.
\section{Experiment}

\subsection{Datasets}

\begin{table*}[t!]
\centering
\scalebox{0.9}{
\setlength\tabcolsep{1.5pt}
\def\arraystretch{1.1}
\begin{tabular}{p{2.5cm}<{\centering} p{0.2cm}<{\centering}p{1.1cm}<{\centering}p{1.1cm}<{\centering}p{1.1cm}<{\centering}p{1.1cm}<{\centering}p{1.1cm}<{\centering}p{1.1cm}<{\centering} p{1.1cm}<{\centering}p{1.1cm}<{\centering}p{1.1cm}<{\centering} p{1.1cm}<{\centering}p{1.1cm}<{\centering}p{1.1cm}<{\centering}p{1.1cm}}
\hline\hline
    Method & & A$\to$C & A$\to$P & A$\to$R & C$\to$A & C$\to$P & C$\to$R & P$\to$A & P$\to$C & P$\to$R & R$\to$A & R$\to$C & R$\to$P & Avg\\
\hline

    ResNet-50~\cite{he2016deep} & & 34.9 & 50.0 & 58.0  & 37.4 & 41.9 & 46.2 & 38.5 & 31.2 & 60.4 & 53.9 & 41.2 & 59.9 & 46.1\\
    
    DAN~\cite{long2015learning} & & 43.6 & 57.0 & 67.9  &  45.8 & 56.5 & 60.4  & 44.0  & 43.6 & 67.7 & 63.1 & 51.5 & 74.3 & 56.3 \\
    
    DANN~\cite{ganin2016domain} & & 45.6 & 59.3 & 70.1  &  47.0 & 58.5 & 60.9  & 46.1  & 43.7 & 68.5 & 63.2 & 51.8 & 76.8 & 57.6 \\
    
    MCD~\cite{saito2018maximum} & & 48.9 & 68.3 & 74.6  &  61.3 & 67.6 & 68.8  & 57.0  & 47.1 & 75.1 & 69.1 & 52.2 & 79.6 & 64.1 \\
    
    MDD~\cite{zhang2019bridging} & & 54.9 & 73.7 & 77.8  &  60.0 & 71.4 & 71.8  & 61.2  & 53.6 & 78.1 & 72.5 & 60.2 & 82.3 & 68.1 \\
    
    CDAN~\cite{long2018conditional} & & 50.7 & 70.6 & 76.0  &  57.6 & 70.0 & 70.0  & 57.4  & 50.9 & 77.3 & 70.9 & 56.7 & 81.6 & 65.8 \\
    
    SymNets~\cite{zhang2019domain} & & 47.7 & 72.9 & 78.5  &  64.2 & 71.3 & 74.2  & 63.6  & 47.6 & 79.4 & 73.8 & 50.8 & 82.6 & 67.2 \\
    
    CADA~\cite{kurmi2019attending} & & 56.9 & \textbf{76.4} & \textbf{80.7} & 61.3 & \textbf{75.2} & \textbf{75.2} & 63.2 & 54.5 & 80.7 & 73.9 & \textbf{61.5} & 84.1 & 70.2 \\
    
    ETD~\cite{li2020enhanced} & & 51.3 & 71.9 & 85.7 & 57.6 & 69.2 & 73.7 & 57.8 & 51.2 & 79.3 & 70.2 & 57.5 & 82.1 & 67.3\\
    
    GVB-GD~\cite{cui2020gradually} & & 57.0 & 74.7 & 79.8  &  \textbf{64.6} & 74.1 & 74.6  & \textbf{65.2}  & 55.1 & \textbf{81.0} & \textbf{74.6} & 59.7 & 84.3 & 70.4 \\
    
    \hline
    
    Baseline & & 54.3 & 70.1 & 75.2 &  60.4 & 71.5 & 72.3  & 60.1  & 52.1 & 74.2 & 71.5 & 56.2 & 81.3 & 66.6 \\
    
    \textbf{TCM (Ours)} & & \cellcolor{mygray}\textbf{58.6} & \cellcolor{mygray}74.4 & \cellcolor{mygray}79.6 & \cellcolor{mygray}64.5 &  \cellcolor{mygray}74.0 & \cellcolor{mygray}75.1 & \cellcolor{mygray}64.6 & \cellcolor{mygray}\textbf{56.2} & \cellcolor{mygray}80.9 & \cellcolor{mygray}\textbf{74.6} & \cellcolor{mygray}60.7 & \cellcolor{mygray}\textbf{84.7} & \cellcolor{mygray}\textbf{70.7}\\
\hline \hline
\end{tabular}}
\caption{Accuracy (\%) on the Office-Home dataset~\cite{venkateswara2017Deep} with 12 UDA tasks, where all methods are fine-tuned from ResNet-50~\cite{he2016deep} pre-trained on ImageNet~\cite{russakovsky2015imagenet}.}
\label{tab:1}
\end{table*}
\begin{table*}[t!]
\centering
\scalebox{0.9}{
\def\arraystretch{1.1}
\begin{tabular}{p{2.5cm}<{\centering} p{0.1cm}<{\centering}p{1.6cm}<{\centering}p{1.6cm}<{\centering}p{1.6cm}<{\centering}p{1.6cm}<{\centering}p{1.6cm}<{\centering}p{1.6cm}<{\centering}p{1.1cm}}
\hline\hline
    Method & & I $\to$ P & P $\to$ I & I $\to$ C & C $\to$ I & C $\to$ P & P $\to$ C & Avg\\
\hline

    ResNet-50~\cite{he2016deep} & & 74.8 $\pm$ 0.3 & 83.9 $\pm$ 0.1 & 91.5 $\pm$ 0.3 & 78.0 $\pm$ 0.2 & 65.5 $\pm$ 0.3 & 91.2 $\pm$ 0.3 & 80.7\\
    
    DAN~\cite{long2015learning} & & 74.5 $\pm$ 0.4 & 82.2 $\pm$ 0.2 & 92.8 $\pm$ 0.2 & 86.3 $\pm$ 0.4 & 69.2 $\pm$ 0.4 & 89.8 $\pm$ 0.4 & 82.5\\
    
    DANN~\cite{ganin2016domain} & & 75.0 $\pm$ 0.6 & 86.0 $\pm$ 0.3 & 96.2 $\pm$ 0.4 & 87.0 $\pm$ 0.5 & 74.3 $\pm$ 0.5 & 91.5 $\pm$ 0.6 & 85.0\\
    
    RevGrad~\cite{ganin2015unsupervised} & & 75.0 $\pm$ 0.6 & 86.0 $\pm$ 0.3 & 96.2 $\pm$ 0.4 & 87.0 $\pm$ 0.5 & 74.3 $\pm$ 0.5 & 91.5 $\pm$ 0.6 & 85.0\\
    
    MADA~\cite{pei2018multi} & & 75.0 $\pm$ 0.3 & 87.9 $\pm$ 0.2 & 96.0 $\pm$ 0.3 & 88.8 $\pm$ 0.3 & 75.2 $\pm$ 0.2 & 92.2 $\pm$ 0.3 & 85.8\\
    
    CDAN~\cite{long2018conditional} & & 77.7 $\pm$ 0.3 & 90.7 $\pm$ 0.2 & 97.7 $\pm$ 0.3 & 91.3 $\pm$ 0.3 & 74.2 $\pm$ 0.2 & 94.3 $\pm$ 0.3 & 87.7\\
    
    A$^2$LP~\cite{zhang2020label} & & 79.6 $\pm$ 0.3 & 92.7 $\pm$ 0.3 & 96.7 $\pm$ 0.1 & 92.5 $\pm$ 0.2 & 78.9 $\pm$ 0.2 & 96.0 $\pm$ 0.1 & 89.4\\
    
    SymNets~\cite{zhang2019domain} & & \textbf{80.2} $\pm$ 0.3 & 93.6 $\pm$ 0.2 & 97.0 $\pm$ 0.3 & 93.4 $\pm$ 0.3 & 78.7 $\pm$ 0.3 & 96.4 $\pm$ 0.1 & 89.9\\
    
    ETD~\cite{li2020enhanced} & & 81.0 & 91.7 & \textbf{97.9} & 93.3 & 79.5 & 95.0 & 89.7 \\
    
    \hline
    
    Baseline & & 77.2 $\pm$ 0.5 & 89.7 $\pm$ 0.2 & 96.1 $\pm$ 0.3 & 92.1 $\pm$ 0.4 & 75.2 $\pm$ 0.3 & 93.5 $\pm$ 0.4 & 87.3\\
    
    \textbf{TCM (Ours)} & & \cellcolor{mygray}79.9 $\pm$ 0.4 & \cellcolor{mygray}\textbf{94.2} $\pm$ 0.2 & \cellcolor{mygray}97.8 $\pm$ 0.3 & \cellcolor{mygray}\textbf{93.8} $\pm$ 0.4 &  \cellcolor{mygray}\textbf{79.9} $\pm$ 0.4 & \cellcolor{mygray}\textbf{96.9} $\pm$ 0.4 & \cellcolor{mygray}\textbf{90.5}\\
\hline \hline
\end{tabular}}
\caption{Accuracy (\%) and the standard deviation on the ImageCLEF-DA dataset~\cite{imageclef2014} with 6 UDA tasks, where all methods are fine-tuned from ResNet-50~\cite{he2016deep} pre-trained on ImageNet~\cite{russakovsky2015imagenet}.}
\label{tab:2}
\end{table*}

We validated TCM on three standard benchmarks for visual domain adaptation:

\noindent\textbf{ImageCLEF-DA}~\cite{imageclef2014} is a benchmark dataset for ImageCLEF 2014 domain adaptation challenge, which contains three domains: 1) Caltech-256
(C), 2) ImageNet ILSVRC 2012 (I) and 3) Pascal VOC 2012 (P).
For each domain, there are 12 categories and 50 images in
each category. We permuted all the three domains and built six transfer tasks, \ie, I$\to$P, P$\to$I, I$\to$C, C$\to$I, C$\to$P, P$\to$C.

\noindent\textbf{Office-Home}~\cite{venkateswara2017Deep} is a very challenging dataset for UDA with 4 significantly different
domains: Artistic images (A), Clipart (C), Product images (P) and Real-World images (R). It contains 15,500 images from 65 categories of everyday objects in the office and home scenes. We evaluated TCM in all 12 permutations of domain adaptation tasks.

\noindent\textbf{VisDA-2017}~\cite{peng2017visda} is a challenging simulation-to-real dataset that is significantly larger than the other two datasets with 280k images in 12 categories. It has two domains: Synthetic, with renderings of 3D models from different angles and with different lighting conditions; and Real with natural real-world images. We followed the common protocol~\cite{long2018conditional,cui2020gradually} to evaluate on Synthetic$\to$Real task.

\subsection{Setup}

\noindent\textbf{Evaluation Protocol}. We followed the common evaluation protocol for UDA~\cite{long2018conditional,zhang2019domain,cui2020gradually}, where all labeled source samples and unlabeled target samples are used to train the model, and the average classification accuracy is compared in each dataset based on three random experiments. Following~\cite{zhang2019domain,zhang2020label}, we reported the standard deviation on ImageCLEF-DA. For fair comparison, our TCM and all comparative methods used the backbone ResNet-50~\cite{he2016deep} pre-trained on ImageNet~\cite{russakovsky2015imagenet}.

\noindent\textbf{Implementation Details}. We implemented each $M_i,M_i^{-1}$ in DCMs $\{(M_i,M_i^{-1})\}_{i=1}^k$ as an encoder-decoder network, consisting of 2 down-sampling convolutional layers, followed by 2 ResNet blocks and 2 up-sampling convolutional layers. The loss for training DCMs $\mathcal{L}^i_{CycleGAN}$ consists of the adversarial loss, cycle consistency loss, and identity loss following the official code of CycleGAN. The encoder and decoder network in VAE were each implemented with 2 fully-connected layers. The min-max objective in Eq.~\eqref{eq:proxy_objective} for the proxy loss was implemented using the gradient reversal layer~\cite{ganin2015unsupervised}. The number of DCMs $k$ is a hyperparameter. We used $k=4$ for all Office-Home experiments and the VisDA-2017 experiment and conducted ablation on $k$ with ImageCLEF-DA. 

\noindent\textbf{Baseline}. As existing domain-mapping methods either focus on the image segmentation task~\cite{Li_2019_CVPR}, or only evaluate on toy settings (\eg, digit)~\cite{hoffman2018cycada}, we implemented a domain-mapping baseline. Specifically, we trained a CycleGAN that transforms $X_s\to X_t$ and $X_t\to X_s$, whose network architecture is the same as each DCM in our proposed TCM. Then we learned a classifier on the transformed samples $X_s\to X_t$ with the standard cross-entropy loss, while using the loss in Eq.~\eqref{eq:la} to align the features of the transformed samples with the target domain features. In inference, we directly used the learned classifier.

\subsection{Results}

\begin{figure}[t!]
\centering

\begin{minipage}[b]{.5\linewidth}
    \centering
    \captionsetup{font=footnotesize,labelfont=footnotesize,skip=5pt,width=0.9\linewidth}
    \scalebox{0.9}{
\def\arraystretch{1.1}
\begin{tabular}{p{2.2cm}<{\centering} p{1.1cm}<{\centering}}
\hline\hline
    Method & Acc.\\
\hline
    DAN~\cite{long2015learning} & 61.6\\
    
    DANN~\cite{ganin2016domain} & 57.4\\
    
    GTA~\cite{sankaranarayanan2018generate} & 69.5\\
    
    MDD~\cite{zhang2019bridging} & 74.6\\
    
    CDAN~\cite{long2018conditional} & 70.0\\
    
    GVB-GD~\cite{cui2020gradually} & 75.3\\
    
    DMRL~\cite{wu2020dual} & 75.5\\
    
    \hline
    
    Baseline & 72.8\\
    
    \textbf{TCM (Ours)} & \cellcolor{mygray}\textbf{75.8}\\
\hline\hline
\end{tabular}}
    \captionof{table}{Accuracy (\%) of Synthetic$\to$Real task on the VisDA-2017 dataset ~\cite{peng2017visda}.}
    \label{tab:3}
\end{minipage}%
\begin{minipage}[b]{.5\linewidth}
    \centering
    \captionsetup{font=footnotesize,labelfont=footnotesize,skip=5pt}
    \includegraphics[width=0.9\linewidth]{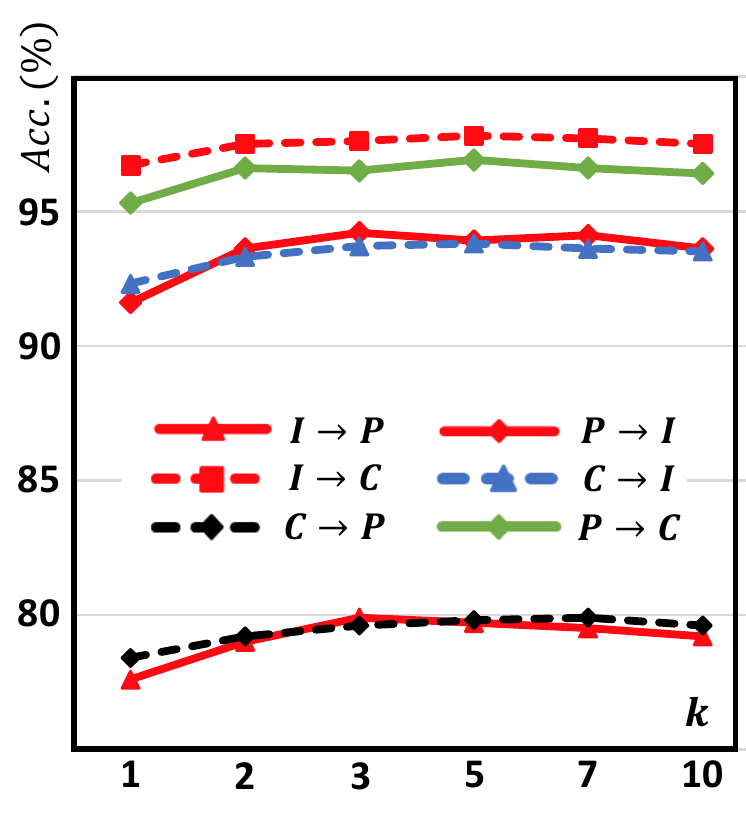}
    \caption{Plot of $k$ against Accuracy (\%) on the 6 UDA tasks from ImageCLEF-DA~\cite{imageclef2014}.}
    \label{fig:6}
\end{minipage}

\end{figure}

\noindent\textbf{Overall Results}. As shown in Table~\ref{tab:1},~\ref{tab:2},~\ref{tab:3}, our method achieves the state-of-the-art average classification accuracy on Office-Home~\cite{venkateswara2017Deep}, ImageCLEF-DA~\cite{imageclef2014}, and VisDA-2017~\cite{peng2017visda}. Specifically, 1) ImageCLEF-DA has the smallest domain gap, where the 3 domains correspond to real-world images from 3 datasets. Our TCM outperforms existing methods on 4 out of 6 UDA tasks.
2) Office-Home has a much larger domain gap, such as the Artistic domain (A) and Clipart domain (C). Besides improvements in average accuracy, our method significantly outperforms existing methods on the most difficult tasks, \eg, $+1.6\%$ on A$\to$C, $+1.1\%$ on P$\to$C.
3) VisDA-2017 is a large-scale dataset with 280k images, where TCM performs competitively. Note that the training complexity and convergence speed of TCM are comparable with existing methods (details in Appendix). Hence overall, our TCM is generally applicable to small to large-scale UDA tasks with various sizes of domain gaps. 

\noindent\textbf{Comparison with Baseline}.
We have three observations:
1) From Table~\ref{tab:1},~\ref{tab:2},~\ref{tab:3}, we notice that the Baseline does not perform competitively on 3 benchmarks. This shows that the improvements from TCM are not the results of superior image generation quality from CycleGAN.
2) Baseline accuracy in Table~\ref{tab:2} is much lower compared with our TCM with a single DCM ($k=1$) in Table~\ref{tab:4} on ImageCLEF-DA. The only difference between them is that Baseline train a classifier directly with the transformed sample $X_s\to X_t$, while TCM($k=1$) uses the proxy function $h_y(X,\hat{X})$ in Eq.~\eqref{eq:practical_transport}. Hence this validates that the proxy theory in Section~\ref{sec:3.1} has some effects in removing the confounding effect, even without stratifying $U$. As shown later, this is because $\hat{X}$ can make up for the lost semantic in $X$.
3) On all 3 benchmarks, when using multiple DCMs, TCM significantly outperforms Baseline, which validates our practical implementation of Eq.~\eqref{eq:practical_transport} by combining DCMs and proxy theory to identify the stratification and representation of $U$.

\begin{table}[t!]
\centering
\scalebox{0.9}{
\setlength\tabcolsep{1.5pt}
\def\arraystretch{1.1}
\begin{tabular}{p{0.3cm}<{\centering}p{1.1cm}<{\centering}p{1.1cm}<{\centering}p{1.1cm}<{\centering}p{1.1cm}<{\centering}p{1.1cm}<{\centering}p{1.1cm}<{\centering}p{1.1cm}}
\hline\hline
    $k$ & I $\to$ P & P $\to$ I & I $\to$ C & C $\to$ I & C $\to$ P & P $\to$ C & Avg\\
\hline

    1 & 77.6 & 91.6 & 96.7 &92.3 & 78.4 & 95.3 & 88.7\\
    
    2 & 79 & 93.6 & 97.5 & 93.3 & 79.2 & 96.6 & 89.9\\
    
    3 & \textbf{79.9} & \textbf{94.2} & 97.6 & 93.7 & 79.6 & 96.5 & 90.3\\
    
    5 & 79.7 & 93.9 & \textbf{97.8} & \textbf{93.8}	& 79.8 & \textbf{96.9} & 90.3\\
    
    7 & 79.5 & 94.1 & 97.7 & 93.6 & \textbf{79.9} & 96.6 & 90.2\\
    
    10 & 79.2 & 93.6 & 97.5 & 93.5 & 79.6 & 96.4 & 90.0\\

\hline \hline
\end{tabular}}
\caption{Ablation on the number of DCMs $k$ with accuracy (\%) on the 6 UDA tasks from ImageCLEF-DA~\cite{imageclef2014}.}
\label{tab:4}
\end{table}
\begin{figure}[t]
    \centering
    \includegraphics[width=.95\linewidth]{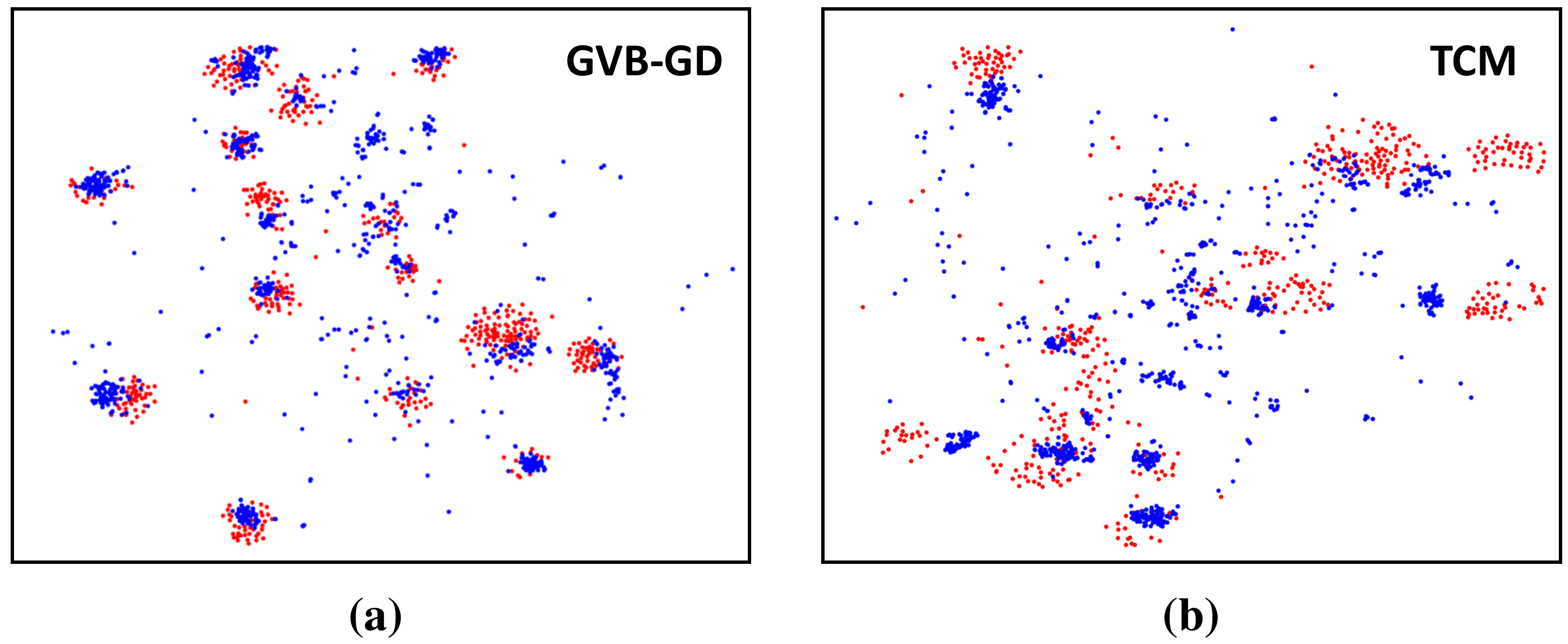}
    \caption{t-SNE~\cite{maaten2008visualizing} plot of features in $S=s$ (red) and in $S=t$ (blue) after training using method: (a) GVB-GD~\cite{cui2020gradually}, (b) Our TCM. Only samples from the first 15 classes are plotted to avoid clutter.}
    \label{fig:7}
\end{figure}

\begin{figure*}[t]
    \centering
    \includegraphics[width=1\linewidth]{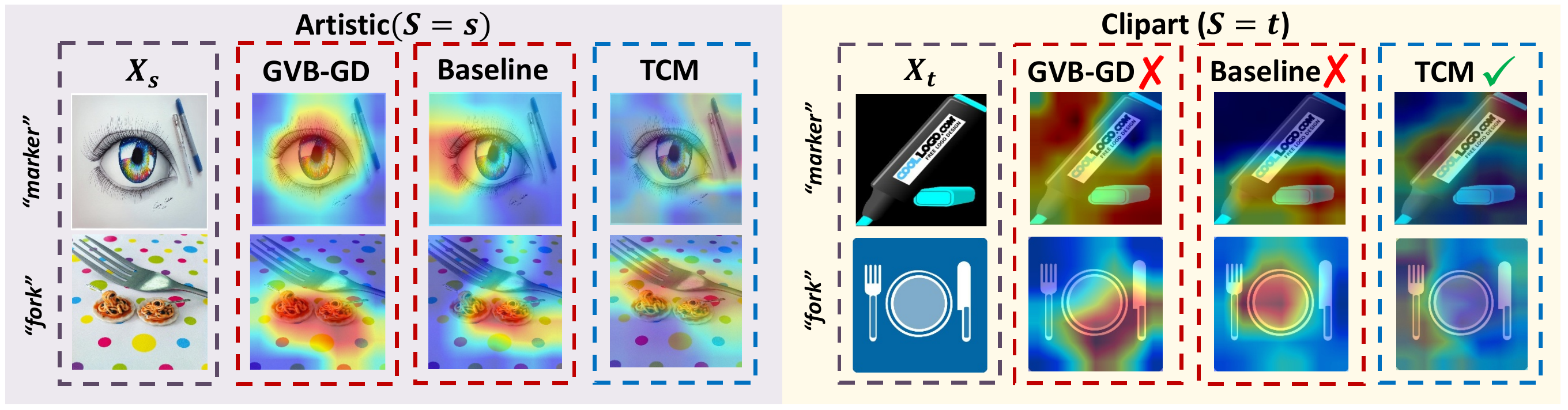}
    \caption{Class Activation Maps (CAMs)~\cite{zhou2016learning} of GVB-GD~\cite{cui2020gradually}, Baseline and TCM on the Artistic$(S=s)\to$Clipart$(S=t)$ task in Office-Home dataset~\cite{venkateswara2017Deep}. The left column shows two samples in each domain, whose class name is indicated on the left. In $S=t$, we show two samples predicted wrongly by GVB-GD and Baseline, but correctly with TCM.}
    \label{fig:8}
    \vspace*{-4mm}
\end{figure*}

\noindent\textbf{Number of DCMs}. We performed an ablation on the number of DCMs $k$ with ImageCLEF-DA dataset. The results are shown in Table~\ref{tab:4} and plotted in Figure~\ref{fig:6}. We have three observations:
1) Among each step of $k$, the largest difference occurs between $k=1$ and $k=2$. This means that accounting for 2 disentangled factors can explain much of the domain shift due to $P(U|S=s)\to P(U|S=t)$ in ImageCLEF-DA.
2) Across all tasks, the best performance is usually achieved with $k=5$. This supports the effectiveness of stratifying $U$ according to a small number of disentangled attributes.
3) A large $U$ (\eg, 10) leads to a slight performance drop. In practice, we observed that when $k$ is large, a few DCMs seldom win any sample from Eq.~\eqref{eq:dcm_objective}, and hence are poorly trained. This can impact their image generation quality, leading to reduced performance. We argue that this is because the chosen $k$ is larger than the number of factors that can be disentangled from the dataset. For example, if there exist $k'<k$ factors $(U_1,\ldots,U_k')$, once $k'$ DCMs establish correspondence with them, the rest $(k-k')$ DCMs will never win any sample, as they can never generate images that are more counterfactual faithful compared with intervention on each $U_i$ (see Counterfactual Faithfulness theorem in Appendix).

\noindent\textbf{t-SNE Plot}.
In Figure~\ref{fig:7}, we show the t-SNE~\cite{maaten2008visualizing} plots of the features in $S=s$ and $S=t$ in Office-Home A$\to$C task after training, for both the current SOTA GVB-GD~\cite{cui2020gradually} and our TCM. GVB-GD is based on invariant-feature-learning and focuses on making $X_s$ and $X_t$ alike, while our TCM does not explicitly align them. We indeed observe a better alignment between the features in $S=s$ and $S=t$ in GVB-GD. However, as shown in Table~\ref{tab:1}, our method outperforms GVB-GD on A$\to$C. This shows that the current common view on the solution to UDA based on~\cite{ben2010theory}, \ie, making domains alike while minimizing source risk, does not necessarily lead to the best performance. In fact, this is in line with the conclusion from recent theoretical works~\cite{zhao2019on}. Our approach is orthogonal to existing approaches and we demonstrate the practicality of a more principled solution to UDA, \ie, establishing appropriate causal assumption on the domain shift and solving the corresponding transport formula (\eg, Eq.~\eqref{eq:transportability}).

\noindent\textbf{Alleviating Semantic Loss}. We use Figure~\ref{fig:8} to reveal the semantic loss problem on existing methods, and how our method can alleviate it. The figure shows the CAM~\cite{zhou2016learning} responses on images in $S=s$ and $S=t$ from the $A\to C$ task in Office-Home dataset.
We compared with two lines of existing approaches: invariant-feature-learning method (GVB-GD~\cite{cui2020gradually}) and domain-mapping method (our Baseline), and we generated CAM for each of them as shown in the red dotted box.
For our TCM, the prediction is a linear combination of the effects from the two images: $X$ and the DCMs outputs $\hat{X}$ (see Eq.~\eqref{eq:h_y}). As the locations of objects tend to remain the same in $\hat{X}$ as in $X$ (see Figure~\ref{fig:4}), we visualize the overall CAM of our TCM in the blue box by combining the CAM from $X$ and $\hat{X}$ weighted by their contributions towards softmax prediction probability.
On the left, we show the CAMs in $S=s$. By looking at the CAM of GVB-GD and the Baseline, we observe that they focus on the contextual object semantics (\eg, food that commonly appears together with ``fork'') to distinguish ``marker'' and ``fork'', where the effect from the object shape is mostly lost. In contrast, our TCM focuses on the marker and fork, \ie, the shape semantic is preserved in training.
On the right in $S=t$, the contextual object semantic is no longer discriminative for the two classes. Hence GVB-GD and Baseline either focus on the wrong semantic (\eg, plate) or become confused (focusing on a large area), leading to the wrong prediction. Thanks to the preserved shape semantic, our TCM focuses on the object and makes the correct prediction. This provides an intuitive explanation of how our TCM alleviates the semantic loss (more examples in Appendix).
\section{Conclusion}

We presented a practical implementation of the transportability theory for UDA called Transporting Causal Mechanisms (TCM). It
systematically eliminates the semantic loss from the confounding effect. In particular, we proposed to identify the confounder stratification by discovering disentangled causal mechanisms and represent the unknown confounder representation by proxy variables. Extensive results on UDA benchmarks showed that TCM is empirically effective. It is worth highlighting that TCM is orthogonal to the existing UDA solutions---instead of making the two domains similar, we aim to find how to transport and what can be transported with a causality theoretic viewpoint of the domain shift. We believe that the key bottleneck of TCM is the quality of causal mechanism disentanglement. Therefore, we will seek more effective unsupervised feature disentanglement methods and investigate their causal mechanisms.
\section{Acknowledgements}

The authors would like to thank all reviewers and ACs for their constructive suggestions, and specially thank Alibaba City Brain Group for the donations of GPUs. This research is partly supported by the Alibaba-NTU Singapore Joint Research Institute, Nanyang Technological University (NTU), Singapore; A*STAR under its AME YIRG Grant (Project No. A20E6c0101); and the Singapore Ministry of Education (MOE) Academic Research Fund (AcRF) Tier 2 grant. 
\clearpage
\renewcommand{\thesection}{A.\arabic{section}}
\renewcommand*{\thesubsection}{A.\arabic{section}.\arabic{subsection}}
\renewcommand{\thetable}{A\arabic{table}}
\renewcommand{\thefigure}{A\arabic{figure}}
\setcounter{section}{0}
\setcounter{figure}{0}
\setcounter{table}{0}

\noindent\textbf{\Large Appendix}
\vspace{0.1in}

This appendix is organized as follows:

\begin{itemize}
    \item For preliminaries on structural causal model and do-calculus, we refer readers to Section 2 of~\cite{pearl2014external}.
    \item Section~\ref{sec:a1} gives the proofs and derivations for Section 3, where we first prove the Counterfactual Faithfulness theorem in Section~\ref{sec:a11}, and then prove the sufficient condition used to establish the correspondence between DCMs and disentangled generative causal factors in Section~\ref{sec:a12}.
    \item Section~\ref{sec:a2}  the proofs and derivations for Section 4, where we prove the Proxy Function theorem and its corollary in Section~\ref{sec:a21}, and then derive Eq. (7) in Section~\ref{sec:a22}.
    \item Section~\ref{sec:a3} provides implementation details. Specifically, in Section~\ref{sec:a31}, we provide the network architecture of DCMs, the implementation of CycleGAN loss $\mathcal{L}_{CycleGAN}^i$ and DCM training details. In Section~\ref{sec:a32}, we show the network architectures of our backbone, the VAE and discriminators, together with their training details. In Section~\ref{sec:a33}, we attend to some details in the experiment.
    \item Section~\ref{sec:a4} shows additional generated images from our DCMs and additional CAM results.
\end{itemize}

\section{Proof and Derivation for Section 3}
\label{sec:a1}
In this section, we will first derive the \emph{Counterfactual Faithfulness theorem}. Then we will prove the sufficient condition in Section 3.1.

\subsection{Counterfactual Faithfulness Theorem}
\label{sec:a11}

We will first provide a brief introduction to the concept of counterfactual and disentanglement. Causality allows to compute how an outcome would have changed, had some variables taken different
values, referred to as a counterfactual.
In Section 3.1, we refer to each DCM $(M_i,M_i^{-1})$ in $\{(M_i,M_i^{-1})\}_{i=1}^k$ as a counterfactual mapping, where each $M_i$ (or $M_i^{-1}$) essentially follows the three steps of computing counterfactuals~\cite{pearl2018book} (conceptually): given a sample $X=\mathbf{x}$, 1) In abduction, $(U_1=u_1,\ldots,U_i=u_i,\ldots,U_k=u_k)$ is inferred from $\mathbf{x}$ through $P(U|X)$; 2) In action, the attribute $U_i$ is intervened by setting it to $u_i'$ drawn from $P(U_i|S=t)$ (or $P(U_i|S=s)$), while the values of other attributes are fixed; 3) In prediction, the modified $(U_1=u_1,\ldots,U_i=u_i',\ldots,U_k=u_k)$ is fed to the generative process $P(X|U)$ to obtain the output of the DCM $M_i(\mathbf{x})$ (or $M_i^{-1}(\mathbf{x})$). More details regarding counterfactual can be found in~\cite{pearl2016causal}.

Our definition of disentanglement is based on~\cite{higgins2018towards} of group theory. 
Let $\mathcal{U}$ be a set of (unknown) generative factors, \eg, such as shape and background. There is a set of \emph{independent causal mechanisms} $\varphi: \mathcal{U}\to \mathcal{X}$, generating images from $\mathcal{U}$. Let $\mathcal{G}$ be the group acting on $\mathcal{U}$, \ie, $g\circ u$ transforms $u\in \mathcal{U}$ using $g\in \mathcal{G}$ (\eg, changing background ``cluttered'' to ``pure''). When there exists a direct product decomposition $\mathcal{G}=\prod_{i=1}^k \mathcal{G}_i$ and $\mathcal{U}=\prod_{i=1}^k \mathcal{U}_i$ such that $\mathcal{G}_i$ acts on $\mathcal{U}_i$, we say that each $\mathcal{U}_i$ is the space of a disentangled factor. The causal mechanism $(M_i,M_i^{-1})$ is disentangled when its transformation in $\mathcal{X}$ corresponds to the action of $\mathcal{G}_i$ on $\mathcal{U}_i$.


We use $\mathcal{U}$ and $\mathcal{X}$ to denote the vector space of $U$ and $X$ respectively. We denote the generative process $P(X|U)$ ($U\to X$) as a function $g:\mathcal{U}\to \mathcal{X}$. 
Note that we consider the function $g$ as an embedded function~\cite{besserve2020counterfactual}, \ie, a continuous injective function with continuous inversion, which generally holds for convolution-based networks as shown in~\cite{puthawala2020globally}.
Without loss of generality, we will consider the $S=s\to S=t$ mapping $M_i$ for the analysis below, which can be easily extended to $M_i^{-1}$. Our definition of disentangled intervention follows the intrinsic disentanglement definition in~\cite{besserve2020counterfactual}, given by:

\noindent\textbf{Definition} (Disentangled Intervention). \textit{A counterfactual mapping $M:\mathcal{X}\to\mathcal{X}$ is a disentangled intervention with respect to $U_i$, if there exists a transformation $M':\mathcal{U}\to \mathcal{U}$ affecting only $U_i$, such that for any $u\in \mathcal{U}$,}
\begin{equation}
    M(g(u))=g(M'(u)).
\end{equation}

Then we have the following theorem:

\noindent\textbf{Theorem} (Counterfactual Faithfulness Theorem). \emph{The counterfactual mapping $M_i(X)$ is faithful if and only if $M_i$ is a disentangled intervention with respect to $U_i$.}

Note that by definition, if $M_i(X)$ is faithful, $M_i(X)\in \mathcal{X}$.
To prove the above theorem, one conditional is trivial: if
$M_i$ is a disentangled intervention, it is by definition an endomorphism of $\mathcal{X}$ so the counterfactual mapping must be faithful. For the second conditional, let us assume a faithful counterfactual mapping $M_i(X)$. Given $g$ is embedded, the counterfactual mapping can be decomposed as:
\begin{equation}
    M_i(X)=g \circ M_i' \circ g^{-1}(X),
\end{equation}
where $\circ$ denotes function composition and $M'_i:\mathcal{U}\to \mathcal{U}$ affecting only $U_i$. Now for any $u\in \mathcal{U}$, the quantity $M_i(g(u))$ can be similarly decomposed as:
\begin{equation}
    M_i(g(u)) = g \circ M_i' \circ g^{-1} \circ g(u) = g \circ M_i'(u).
\end{equation}
Since $M_i'$ is a transformation in $\mathcal{U}$ that only affects $U_i$, we show that faithful counterfactual transformation $M_i(X)$ is a disentangled intervention with respect to $U_i$, hence completing the proof.

With this theory, faithfulness$=$disentangled intervention. In Section 3.1, we train $M_i$ to ensure $M_i(\mathbf{x}_s)\sim P(X_t)$ (faithfulness) for every sample $\mathbf{x}_s$ in $S=s$, hence encouraging $M_i$ to be a disentangled intervention. Note that the above analysis can easily generalize to $M_i^{-1}$.

\subsection{Sufficient Condition}
\label{sec:a12}

We will prove the following sufficient condition: \emph{if $(M_i,M_i^{-1})$ intervenes $U_i$, the $i$-th mapping function outputs the counterfactual faithful generation, \ie, the smallest $\mathcal{L}_{CycleGAN}^i$}.

Without loss of generality, we will prove for the $S=s \to S=t$ mapping $M_i$, which can be extended to $M_i^{-1}$. For a sample $\mathbf{x}_s$ in $S=s$, let $g^{-1}(\mathbf{x}_s)=\mathbf{u}=(u_1,\ldots,u_i,\ldots,u_k)$. We modify $U_i$ by changing $u_i$ to a value $\hat{u}_i$ drawn from $P(U_i|S=t)$. Denote the modified attribute as $\hat{\mathbf{u}}=(u_1,\ldots,\hat{u}_i,\ldots,u_k)$. Denote the sample with attribute $\hat{\mathbf{u}}$ as $\hat{\mathbf{x}}$. Given $M_i$ intervenes $U_i$, $M_i(\mathbf{x}_s)$ corresponds to a counterfactual outcome when $U_i$ is set to $\hat{u}_i$ through intervention (or $U$ set as $\hat{\mathbf{u}}$). Now as $g^{-1}(\hat{\mathbf{x}})=\hat{\mathbf{u}}$, using the \emph{counterfactual consistency rule}~\cite{pearl2009causality}, we have $M_i(\mathbf{x}_s)=\hat{\mathbf{x}}$. As $\hat{\mathbf{x}}$ is faithful with the Counterfactual Faithfulness theorem, we prove that $M_i(\mathbf{x}_s)$, \ie, the output of the $i$-th mapping function, is also faithful,\ie, the smallest $\mathcal{L}_{CycleGAN}^i$.

\section{Proof and Derivation for Section 4}
\label{sec:a2}

In this section, we will first derive the Proxy Function theorem and the domain-agnostic nature of the proxy function, and then derive Eq. (7) under our chosen function forms in Section 4.

\subsection{Proxy Function Theorem}
\label{sec:a21}

We will derive for the general case where $\tilde{X}$ is any continuous proxy. We will assume that the confounder $U$ follows the completeness condition in~\cite{miao2018identifying}, which accommodates most commonly-used parametric and semi-parametric models such as exponential families.

Given $h_y(X,\hat{X})$ solves Eq. (3), we have:
\begin{equation}
\begin{split}
    &P(Y|Z,X,S) = \int_{-\infty}^{+\infty} h_y(X,\hat{X}) P(\hat{X}|Z,X,S) d\hat{X}\\
    &= \int_{-\infty}^{+\infty} h_y(X,\hat{X}) \left\{\int_{-\infty}^{+\infty} P(\hat{X}|U) P(U|Z,X,S) dU \right\} d\hat{X}.
    \label{eq:proxya1}
\end{split}
\end{equation}

From the law of total probability, we have:
\begin{equation}
    P(Y|Z,X,S) = \int_{-\infty}^{+\infty} P(Y|U,X) P(U|Z,X,S) dU.
    \label{eq:proxya2}
\end{equation}
With Eq.~\eqref{eq:proxya1} and Eq.~\eqref{eq:proxya2} and the completeness condition, we have:
\begin{equation}
    P(Y|U,X)= \int_{-\infty}^{+\infty} h_y(X,\hat{X}) P(\hat{X}|U) d\hat{X},
    \label{eq:proxy3}
\end{equation}
which proves the Proxy Function theorem.

From Eq.~\eqref{eq:proxy3}, we have:
\begin{equation}
\begin{split}
    &\int_{-\infty}^{+\infty} h_y(X,\hat{X},S=s) P(\hat{X}|U) d\hat{X} \\
    &= \int_{-\infty}^{+\infty} h_y(X,\hat{X},S=t) P(\hat{X}|U) d\hat{X}.
\end{split}
\end{equation}
Hence from the completeness condition, we have $h_y(X,\hat{X},S=s)=h_y(X,\hat{X},S=t)$~\cite{miao2018identifying}. Hence we prove that $h_y(X,\hat{X})$ is domain-agnostic.

Note that in Section 4, our proxy $\hat{X}$ is a continuous random variable takes values from $\{M_i(\mathbf{x}_s)\}_{i=1}^k$ for sample $\mathbf{x}_s$ in $S=s$ or $\{M_i^{-1}(\mathbf{x}_t)\}_{i=1}^k$ for sample $\mathbf{x}_t$ in $S=t$. This is a special case of the analysis above with the probability mass of $\hat{X}$ centers around the set of its possible values.

\subsection{Derivation of Eq. (7)}
\label{sec:a22}

We derive Eq. (7) as a corollary to~\cite{miao2018identifying}. The goal is to solve for $h_y(X,\hat{X})$ under the function form in Eq. (6) from the formula below:
\begin{equation}
    P(Y|Z,X,S=s)=\int_{-\infty}^{\infty} P(\hat{X}|Z,X) h_y(X,\hat{X}) d\hat{X}.
\end{equation}

For simplicity, we define a standard multivariate Gaussian function $\phi(\cdot)$. If $A\sim\mathcal{N}(0,\mathbf{I})$ and $A\in \mathbb{R}^n$, we have:
\begin{equation}
    P(A)=\phi(A)=\frac{1}{(2\pi)^{n/2}} exp(-\frac{1}{2}A^TA).
\end{equation}
Our function form for $P(Y|Z,X,S=s)$ is given by $\mathcal{N}(\mathbf{b}_1 + \mathbf{W}_1 Z + \mathbf{W}_2 X,\boldsymbol{\Sigma}_1)$
and for $P(\hat{X}|Z,X,S=s)$ is given by $\mathcal{N}(\mathbf{b}_2 + \mathbf{W}_3 Z + \mathbf{W}_4 X,\boldsymbol{\Sigma}_2)$, where the variance terms are omitted in the main text for brevity, as the final results only depend on the means. Specifically, $\boldsymbol{\Sigma}_2\in \mathbb{R}^{n\times n}$ is a symmetrical matrix with Eigen-decomposition given by $\boldsymbol{\Sigma}_2=\boldsymbol{\mathcal{U}} \boldsymbol{\Lambda} \boldsymbol{\mathcal{U}}^T$, where $\boldsymbol{\mathcal{U}} \in \mathbb{R}^{n\times n}$ is a full-rank matrix containing the eigen-vectors and $\boldsymbol{\Lambda} \in \mathbb{R}^{n\times n}$ is a diagonal matrix with eigen-values. We define $\mathbf{B}=\boldsymbol{\mathcal{U}}\boldsymbol{\Lambda}^{\frac{1}{2}}$, and hence $\boldsymbol{\Sigma}_2=\mathbf{BB}^T$. We can rewrite $P(\hat{X}|Z,X,S=s)$ as:
\begin{equation}
    P(\hat{X}|Z,X,S=s)=|\mathbf{BB}^T|^{-\frac{1}{2}} \phi(\mathbf{B}^{-1}(\hat{X}-\boldsymbol{\hat{\mu}})),
\end{equation}
where $\hat{\mu}=\mathbf{b}_2 + \mathbf{W}_3 Z + \mathbf{W}_4 X$. Define $Z'=\mathbf{B}^{-1}\boldsymbol{\hat{\mu}}$ and $\hat{X}'=\mathbf{B}^{-1}\hat{X}$. We define
\begin{equation}
    t_y(Z',X)=P(Y|Z=\mathbf{W}_3^+ (\mathbf{B}Z' - \mathbf{W}_4 X - \mathbf{b}_1), X, S=s).
\end{equation}
Now we can solve $h_y$ from
\begin{equation}
    t_y(Z',X) = \int_{-\infty}^{\infty} \phi(\hat{X}'-Z') h_y(X,\mathbf{B} \hat{X}') d\hat{X}'.
    \label{eq:proxy_ty}
\end{equation}
Specifically, let $h_1$ and $h_2$ represent the Fourier transform of $\phi$ and $t_y$, respectively:
\begin{equation}
\begin{split}
    h_1(\boldsymbol{\nu})&=\int_{-\infty}^{\infty} exp(-i\boldsymbol{\nu}Z')\phi(Z')dZ'\\
    &=\int_{-\infty}^{\infty} exp(-i\boldsymbol{\nu}Z)\phi(Z)dZ
\end{split}
\label{eq:proxy_h1}
\end{equation}
\begin{equation}
    h_2(\boldsymbol{\nu},X,Y)=\int_{-\infty}^{\infty}exp(-i\boldsymbol{\nu}Z')t_y(Z',X)dZ',
    \label{eq:proxy_h2}
\end{equation}
where $i=(-1)^{\frac{1}{2}}$ is the imaginary unity. Substituting Eq.~\eqref{eq:proxy_ty}~\eqref{eq:proxy_h1} into Eq.~\eqref{eq:proxy_h2}, we have:
\begin{equation}
    h_2(\boldsymbol{\nu},X,Y)=h_1(\boldsymbol{\nu})\int_{-\infty}^{\infty} exp(-i\boldsymbol{\nu}\hat{X}')h_y(X,\mathbf{B} \hat{X}') d\hat{X}'.
\end{equation}
Hence
\begin{equation}
    \int_{-\infty}^{\infty} exp(-i\boldsymbol{\nu}\hat{X}')h_y(X,\mathbf{B} \hat{X}') d\hat{X}'=\frac{h_2(\boldsymbol{\nu},X,Y)}{h_1(\boldsymbol{\nu})}.
\end{equation}
By Fourier inversion, we have:
\begin{equation}
    h_y(X,\mathbf{B} \hat{X}')=\frac{1}{2\pi} \int_{-\infty}^{\infty} exp(-i\boldsymbol{\nu}\hat{X}')\frac{h_2(\boldsymbol{\nu},X,Y)}{h_1(\boldsymbol{\nu})} d\boldsymbol{\nu}
\end{equation}
By substituting $\hat{X}=\mathbf{B}\hat{X}'$, we have
\begin{equation}
    h_y(X,\hat{X})=\frac{1}{2\pi} \int_{-\infty}^{\infty} exp(-i\boldsymbol{\nu}\mathbf{B}^{-1}\hat{X})\frac{h_2(\boldsymbol{\nu},X,Y)}{h_1(\boldsymbol{\nu})} d\boldsymbol{\nu}
    \label{eq:proxy_hy}
\end{equation}
Solving for Eq.~\eqref{eq:proxy_hy} yields:
\begin{equation}
\begin{split}
    h_y(X,\hat{X}) &= \mathbf{b}_1-\mathbf{W}_1\mathbf{W}_3^+\mathbf{b}_2 + \mathbf{W}_1\mathbf{W}_3^+ \hat{X} \\
        &+ (\mathbf{W}_2-\mathbf{W}_1\mathbf{W}_3^+\mathbf{W}_4)X,
\end{split}
\end{equation}
where scaling terms not related with $X,\hat{X}$ are dropped as they do not impact inference. This completes the derivation.

\section{Implementation Details}
\label{sec:a3}

In this section, we will first provide a system overview, followed by the implementation details for DCMs in Section 3 and implementation details for backbone, VAE and discriminators in Section 4. Then we will give a more detailed discussion on the experiment design.

\subsection{Implementation Details for DCMs}
\label{sec:a31}

\begin{figure}[h!]
    \centering
    \includegraphics[width=.85\linewidth]{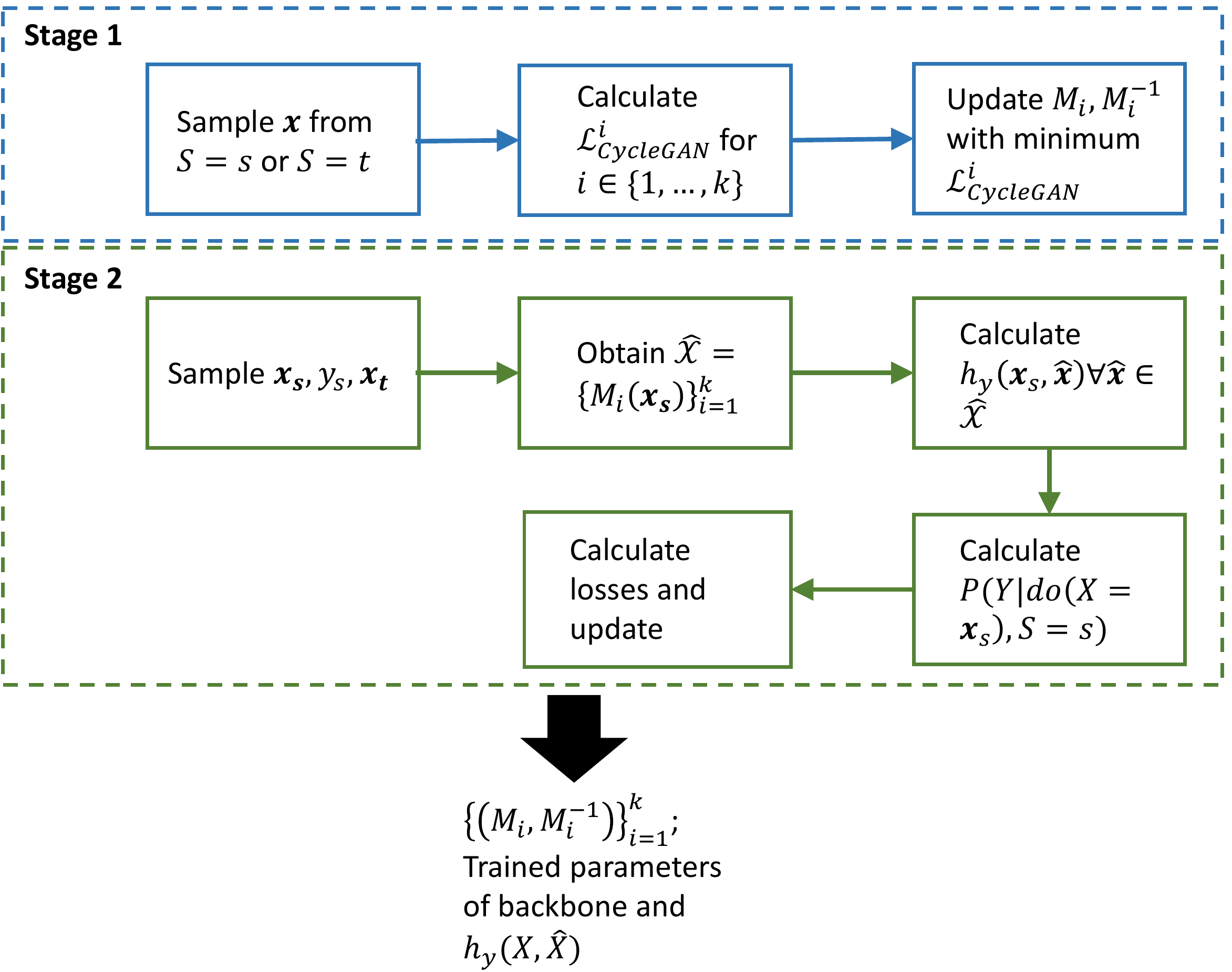}
    \caption{Overview of the two-stage training procedure.}
    \label{fig:overview}
\end{figure}

\noindent\textbf{System Overview}. The two-stage training procedure is depicted in Figure~\ref{fig:overview}. For inference, please refer to Algorithm~\ref{alg:2}.

\noindent\textbf{Network Architecture for DCMs}.
Let \texttt{Conv2D($n$, $c$)} represent 2-d convolutional layer with $n\times n$ as kernel size and $c$ output channels. For Each $M_i,M_i^{-1}$, the network architecture is given by \texttt{ReflectionPad(3) Conv2D(7,64) InstanceNorm ReLU Conv2D(3,128) InstanceNorm ReLU Conv2D(3,256) InstanceNorm ReLU ResNetBlock$\times$2 ConvTranspose2D(3,128) InstanceNorm ReLU ConvTranspose2D(3,64) InstanceNorm ReLU ReflectionPad(3) Conv2D(7,3) Tanh}, where each \texttt{ResNetBlock} is implemented by \texttt{Conv2D(3,256) ReflectionPad(1) Conv2D(3,256)} with skip connection.

\noindent\textbf{Network Architecture for CycleGAN Discriminators}. To calculate each $\mathcal{L}_{CycleGAN}^i$, two discriminators taking image as input is required for $S=s$ and $S=t$, respectively. We implement each discriminator with \texttt{Conv2D(4,64) LeakyReLU(0.2) Conv2D(4,128) InstanceNorm LeakyReLU(0.2) Conv2D(4,256) InstanceNorm LeakyReLU(0.2) Conv2D(4,512) InstanceNorm LeakyReLU(0.2) Conv2D(4,1)}.

\noindent\textbf{CycleGAN Loss}. Next, we will detail the implementation of $\mathcal{L}_{CycleGAN}^i$, which is given by:
\begin{equation}
    \mathcal{L}_{CycleGAN}^i = \mathcal{L}_{adv}^i + \alpha_1 \mathcal{L}_{cyc}^i + \alpha_2 \mathcal{L}_{idt}^i\,
\end{equation}
where $\alpha_1,\alpha_2$ are trade-off parameters. 
The adversarial loss $\mathcal{L}_{adv}^i$ requires the transformed images to look like real images in the opposing domain. For a sample $\mathbf{x}$, $\mathcal{L}_{real}^i$ is given by:
\begin{equation}
    \mathcal{L}_{adv}^i=
    \begin{cases}
    \mathrm{log} (1-D'_t(M_i(\mathbf{x}))),& \text{if } \mathbf{x} \text{ from } S=s,\\
    \mathrm{log}(1-D'_s(M_i^{-1}(\mathbf{x}))),& \text{otherwise,}
    \end{cases}
\end{equation}
where $D'_s,D'_t$ are discriminators that return a large value when its input looks like images in $S=s,S=t$, respectively. $\mathcal{L}_{cyc}^i$ is the cycle consistency loss~\cite{zhu2017unpaired} given by
\begin{equation}
    \mathcal{L}_{cyc}^i=
    \begin{cases}
    \norm{M_i^{-1}(M_i(\mathbf{x}))-\mathbf{x}},& \text{if } \mathbf{x} \text{ from } S=s,\\
    \norm{M_i(M_i^{-1}(\mathbf{x}))-\mathbf{x}},& \text{otherwise,}
    \end{cases}
\end{equation}
where $\norm{\cdot}$ is implemented using L1 norm. We follow CycleGAN~\cite{zhu2017unpaired} to use an identity loss $\mathcal{L}_{idt}^i$ to improve generation quality:
\begin{equation}
    \mathcal{L}_{idt}^i=
    \begin{cases}
    \norm{M_i^{-1}(\mathbf{x})-\mathbf{x}},& \text{if } \mathbf{x} \text{ from } S=s,\\
    \norm{M_i(\mathbf{x})-\mathbf{x}},& \text{otherwise.}
    \end{cases}
\end{equation}

\noindent\textbf{Discriminator Training}. $D'_s,D'_t$ are trained to maximize the following loss:
\begin{equation}
    \mathcal{L}_{d}=
    \begin{cases}
    \mathrm{log} D'_s(\mathbf{x}) + \frac{1}{k} \sum_{i=1}^{k} \mathcal{L}^i_{adv},& \text{if } \mathbf{x} \text{ from } S=s,\\
    \mathrm{log} D'_t(\mathbf{x}) + \frac{1}{k} \sum_{i=1}^{k} \mathcal{L}^i_{adv},& \text{otherwise,}
    \end{cases}
    \label{eq:dis_loss}
\end{equation}
which requires $D'_s,D'_t$ to recognize real samples and reject fake samples generated by \emph{all} DCMs.

\noindent\textbf{Training Details}. The networks are randomly initialized by sampling from normal distribution with standard deviation of 0.02. We use Adam optimizer~\cite{kingma2014adam} with initial learning rate 0.0002, beta1 as 0.5 and beta2 as 0.999. We train the network with the initial learning rate for 100 epochs and decay the learning rate linearly for another 100 epochs on ImageCLEF-DA~\cite{imageclef2014} and OfficeHome~\cite{venkateswara2017Deep}. We train the network with the initial learning rate for 20 epochs and decay the learning rate linearly for another 20 epochs on VisDA-2017~\cite{peng2017visda}. The CycleGAN loss and discriminator loss are updated iteratively. The adopted competitive training scheme to train DCMs can sometimes be sensitive to network initialization. Hence all DCMs are first trained for 8000 iterations, before using the competitive training scheme.

\subsection{Implementation Details for Section 4}
\label{sec:a32}

\begin{wrapfigure}{r}{0.2\textwidth}
    \centering
    \includegraphics[width=.2\textwidth]{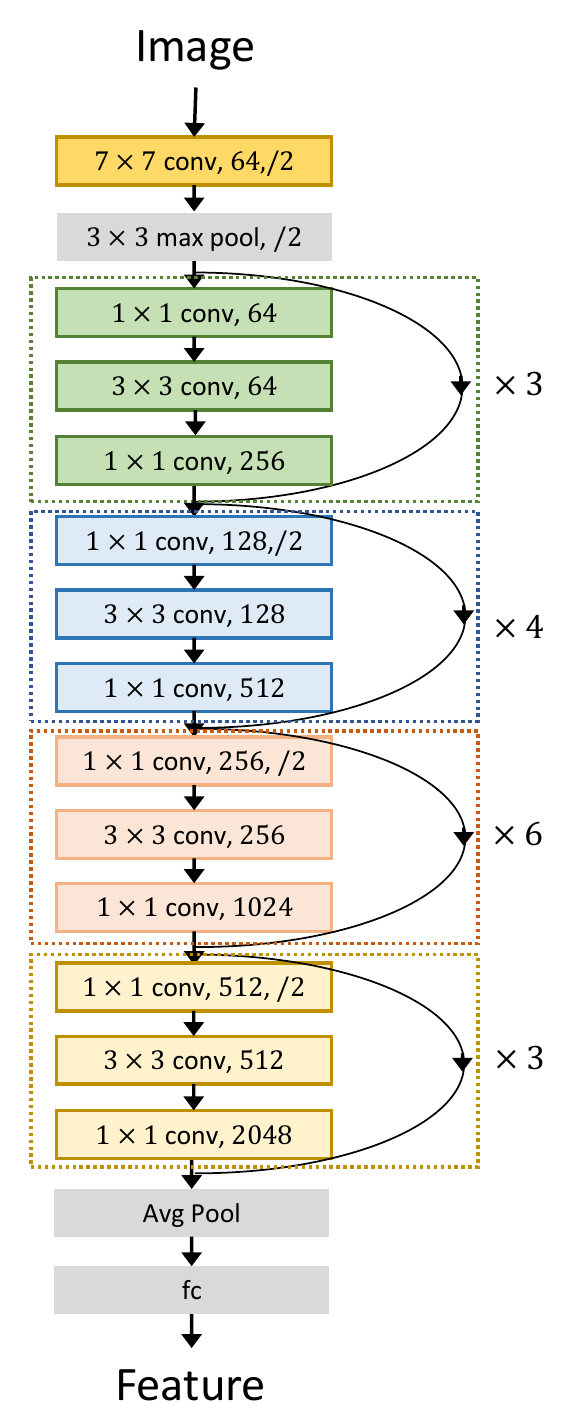}
    \caption{ResNet-50 architecture.}
    \label{fig:resnet50}
\end{wrapfigure}

\noindent\textbf{Backbone}. We adopt ResNet-50~\cite{he2016deep} as our backbone,
where the architecture is shown in Figure~\ref{fig:resnet50}. Specifically, each convolutional layer is described as ($n\times n$ conv, p), where n is the kernel size and p is the number of output channels. Convolutional layers with $/2$ have a stride of 2 and are used to perform down-sampling. The solid curved lines represent skip connection.
The batch normalization and ReLU layers are omitted in Figure~\ref{fig:resnet50}
to highlight the key structure of the backbone. We fine-tuned the pre-trained backbone on ImageNet~\cite{russakovsky2015imagenet} for our experiments.

\noindent\textbf{VAE}. Denote \texttt{Linear($n$,$m$)} as a linear layer with $n$ input channels and $m$ output channels. The encoder $Q_\theta$ is implemented with \texttt{Linear(2048,1200) ReLU Linear(1200,600) ReLU Linear(600,200)}, where the dimension of $Z$ is 100, the first 100 dimensions of the output are the mean and the last 100 dimensions are the variance. The decoder $P_\theta$ is implemented with \texttt{Linear(100,600) ReLU Linear(600,2048) ReLU}.

\noindent\textbf{Discriminators}. The discriminator $D_s$ and $D_t$ are implemented with \texttt{Linear(2048, 1024) ReLU Linear(1024,1024) ReLU Linear(1024,1)}.

\noindent\textbf{Training Details}. The networks are randomly initialized with \texttt{kaiming} initialization with gain as 0.02. We employ mini-batch stochastic gradient descent (SGD) with momentum of 0.9 and nesterov enabled to train our model. We trained the networks for 10000 iterations on VisDA-2017~\cite{peng2017visda} and OfficeHome~\cite{venkateswara2017Deep}, and 3000 iterations on ImageCLEF-DA~\cite{imageclef2014}. The linear functions $f_y,f_{\hat{x}}$, VAE, backbone and discriminators $D_s,D_t$ are updated iteratively.

\subsection{Discussion on Experiment}
\label{sec:a33}

\noindent\textbf{Choice of Dataset}. While Office-31~\cite{saenko2010adapting} is also a popular dataset in UDA, some tasks in the dataset become too trivial, where many UDA algorithms achieve 100\% (or almost) accuracy. Moreover, a recent study~\cite{ringwald2021adaptiope} reveals that the dataset is plagued with wrong labels, ambiguous ground-truth and data leakage, especially in the Amazon domain, which explains the low performance in task DSLR$\to$Amazon and Webcam$\to$Amazon. Hence we followed~\cite{zhang2019domain,long2018conditional,zhang2020label} and performed experiments on ImageCLEF-DA instead, which substitute Office-31 as a small-scale dataset with relatively small domain gap.

\noindent\textbf{Detailed Discussion on t-SNE Plot}. Note that the proxy loss in Section 4 aligns the proxy features with the sample features in the counterpart domain, which is different from existing methods that try to align $X_s$ and $X_t$ directly. Our inference uses the proxy function implemented with Eq. (7), where in training, $X$ takes values of $X_s$ and $\hat{X}$ takes values of $M_i(X_s)$, in testing, $X$ takes values of $X_t$ and $\hat{X}$ takes values of $M_i^{-1}(X_t)$. While $X_s$ and $X_t$ (or $M_i(X_s)$ and $M_i^{-1}(X_t)$) are not aligned in TCM, we can still achieve competitive performance by finding how to transport with a causality theoretic viewpoint of the domain shift.

\section{Additional Results}
\label{sec:a4}

\begin{figure}[h!]
    \centering
    \includegraphics[width=.85\linewidth]{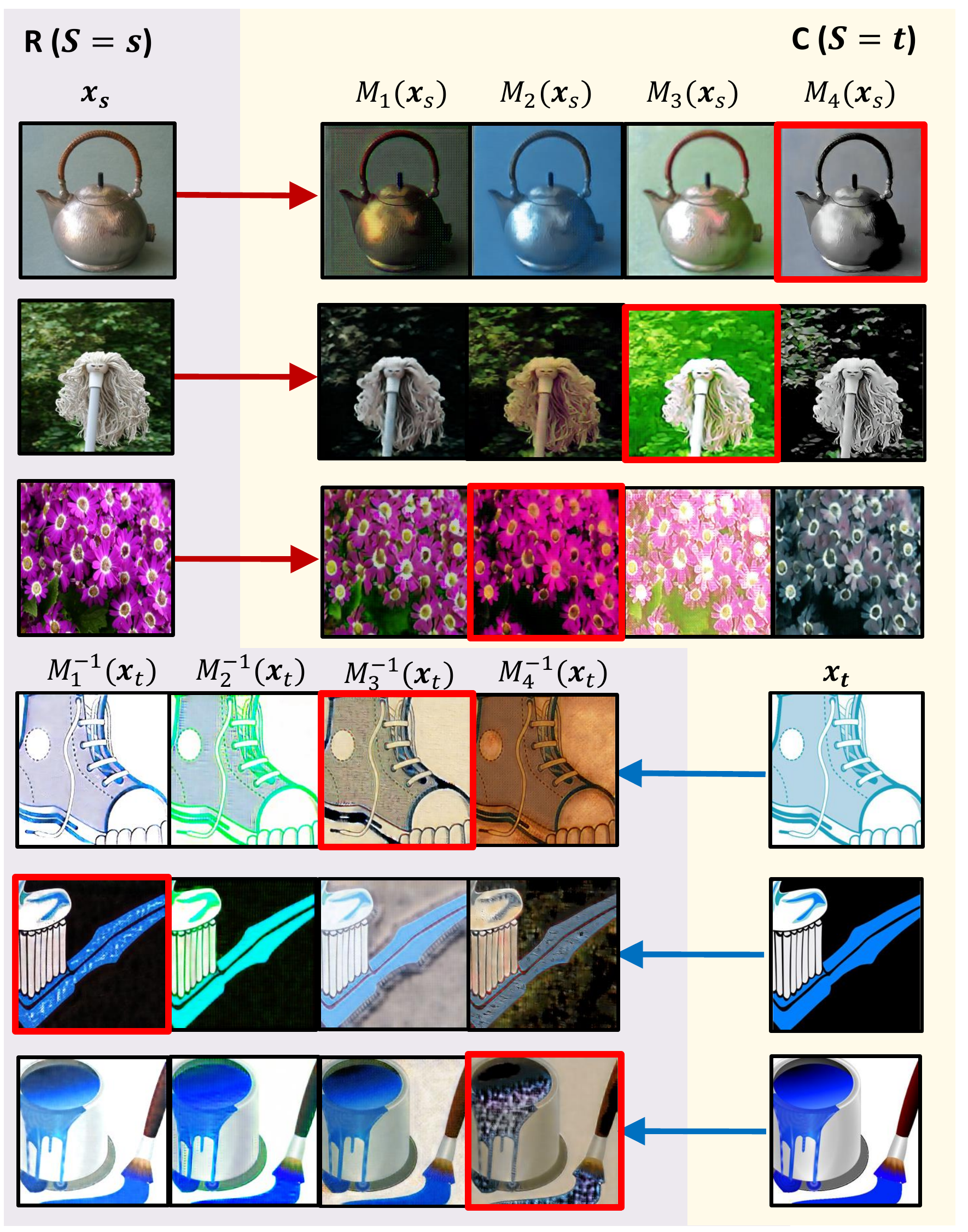}
    \caption{Supplementary to Figure 4.Transformation between ``Real World'' (R) and ``Clipart'' (C) domain with 4 trained DCMs $\{(M_i,M_i^{-1})\}_{i=1}^4$. The winning DCM is outlined in red.}
    \label{fig:visualize}
\end{figure}

\noindent\textbf{Additional Visualizations of DCMs Outputs}. In Figure~\ref{fig:visualize}, we show additional generated images from DCMs in the ``Real World'' (R)$\to$``Clipart'' (C) task of OfficeHome~\cite{venkateswara2017Deep}, which also has a large domain gap. $\{M_1,\ldots,M_4\}$ (or $\{M_1^{-1},\ldots,M_4^{-1}\}$) roughly correspond to reducing (increasing) brightness, changing color, increasing (decreasing) saturation and removing (adding) color.

\begin{figure*}[t]
    \centering
    \includegraphics[width=1\linewidth]{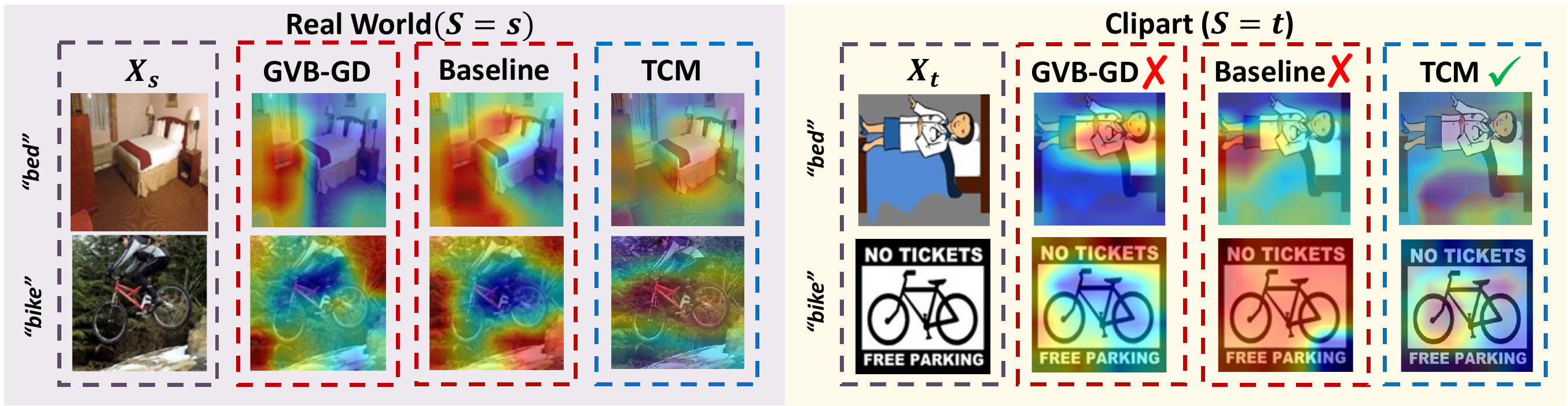}
    \caption{Class Activation Maps (CAMs)~\cite{zhou2016learning} of GVB-GD~\cite{cui2020gradually}, Baseline and TCM on the Real World(S=s)$\to$Clipart$(S=t)$ task in Office-Home dataset~\cite{venkateswara2017Deep}. The left column shows two samples in each domain, whose class name is indicated on the top. In $S=t$, we show two samples predicted wrongly by GVB-GD and Baseline, but correctly with TCM. Supplementary to Figure 8}
    \label{fig:cam_supp}
\end{figure*}

\begin{figure}[h!]
    \centering
    \includegraphics[width=.95\linewidth]{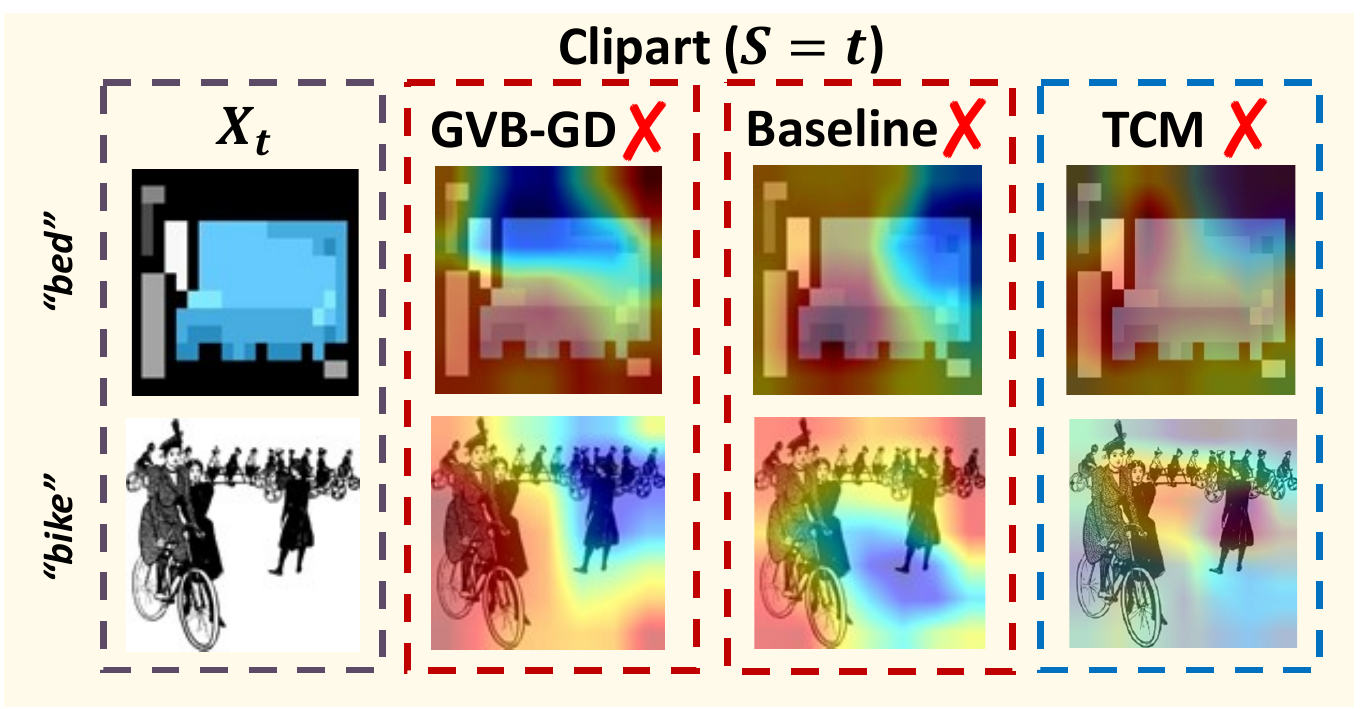}
    \caption{Class Activation Maps (CAMs)~\cite{zhou2016learning} of GVB-GD~\cite{cui2020gradually}, Baseline and TCM on the Real World(S=s)$\to$Clipart$(S=t)$ task in Office-Home dataset~\cite{venkateswara2017Deep}. The left column shows two samples in each domain, whose class name is indicated on the left. All methods fail on the two samples. Supplementary to Figure 8.}
    \label{fig:cam_supp_neg}
\end{figure}

\noindent\textbf{Additional CAM Responses}. In Figure~\ref{fig:cam_supp}, we show the CAM responses in the ``Real World''$\to$``Clipart'' task, where GVB-GD and baseline focuses on the background to distinguish ``bed'' and ``bike'' and the shape semantic is lost, leading to poor generalization in $S=t$. While TCM preserves the foreground object shape semantic. In Figure~\ref{fig:cam_supp_neg}, all three methods fail on the two samples from $S=t$. On the ``bed'' sample, all methods indeed focus on the object. However, the object itself is not very discriminative, which may explain the failure. On the ``bike'' sample, the object is small and all methods fail to distinguish it from the context.

\begin{figure}[t]
    \centering
    \includegraphics[width=1\linewidth]{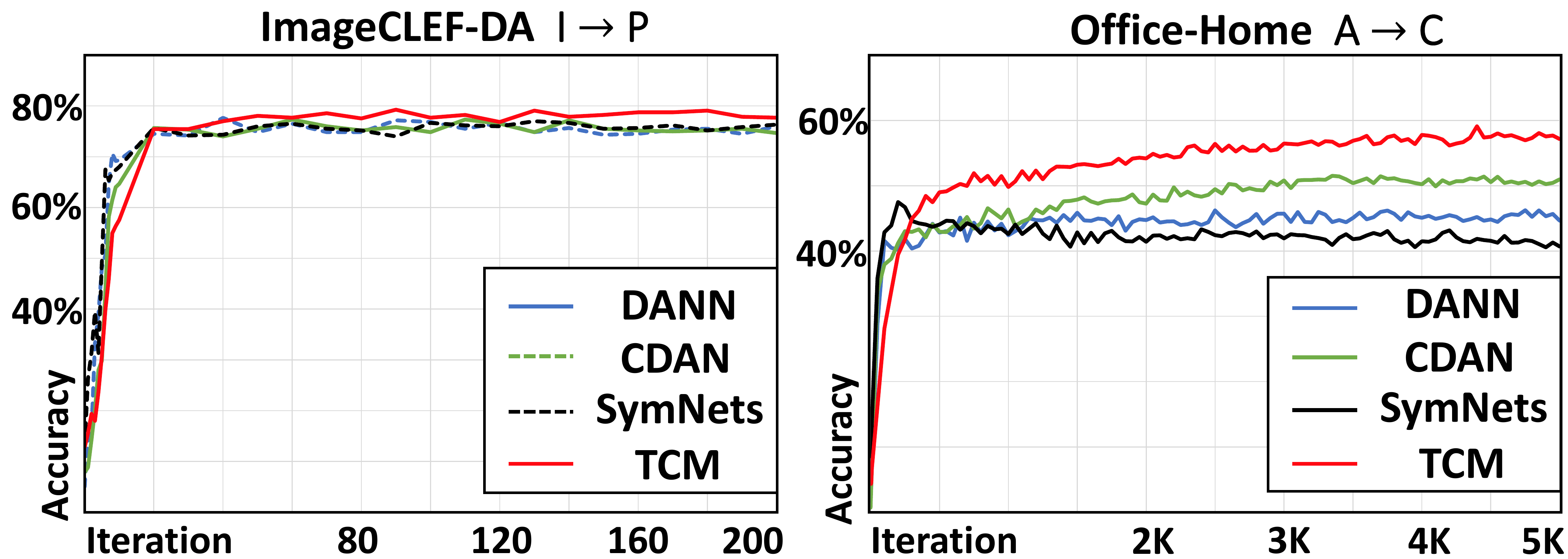}
    \caption{UDA accuracy (\%) using DANN, CDAN, SymNets and our TCM, on different training iterations. Batch size is 32.}
    \label{fig:stats}
\end{figure}

\begin{table}[t!]
\centering
\scalebox{0.9}{
\setlength\tabcolsep{1.5pt}
\begin{tabular}{p{1.7cm}<{\centering}p{2.4cm}<{\centering}p{2.4cm}<{\centering}p{2.0cm}<{\centering}}
\hline\hline
    \scalebox{1.0}{Method} & \scalebox{0.8}{Backbone GFLOPS} & \scalebox{0.8}{Additional GFLOPS} & \scalebox{0.8}{\#Parameters (M)}\\
\hline

    DANN & 133 & 0.084 & 1.3\\
    CDAN & 133 & 1.158 & 18.1\\
    GVB-GD & 133 & 0.092 & 1.4 \\
    Baseline & 133 & 0.072 & 1.1 \\
\hline
    TCM & 133 & 0.364 & 10.1 \\

\hline \hline
\end{tabular}}
\caption{GFLOPS of feature extractor backbone and additional modules (\eg, discriminator networks) as well as \#parameters of those additional modules on Office-Home.}
\vspace{-4mm}
\label{tab:stats}
\end{table}

\noindent\textbf{Convergence Speed, GFLOPS, \#Parameters}. 
The convergence speeds of TCM and related methods are shown in Figure~\ref{fig:stats}. Our TCM (red) converges slightly slower in early iterations (\eg, before 20 on ImageCLEF-DA and 250 on Office-Home) due to the training of VAE in Eq. 9 and the linear layers in Eq. 6. 
After that, TCM converges to the best performance.
The GFLOPS and \#Parameters are shown in Table~\ref{tab:stats}. 
All methods use the same backbone and the backbone fine-tuning takes the major time cost (133), compared with the GFLOPS from additional modules.
%
Regarding the parameters, 
our TCM has more (but not the most) because of the additional VAE and linear layers.

{\small
\bibliographystyle{ieee_fullname}
\bibliography{egbib}
}

\end{document}